\documentclass[11pt,a4paper]{article}
\usepackage[bottom=2.5cm, footskip=1cm]{geometry}
\usepackage{amssymb}
\usepackage{graphicx}
\usepackage{caption}
\usepackage{url}
\usepackage{CJKutf8}
\usepackage[T1]{fontenc}
\usepackage[utf8]{inputenc}
\usepackage{indentfirst} 
\usepackage[backend=bibtex, style=numeric, sorting=none]{biblatex}
\usepackage[ruled,vlined,linesnumbered]{algorithm2e}
\usepackage{booktabs} 
\usepackage{multirow}
\usepackage{amsmath}
\title{FRPSS: Feature Rearrangement in Pre-Shape Space for Single-Image Generation}

\author{
Yuexing Han$^{1,2,*}$ \quad Haoxuan Zhang$^{1}$ \quad Bing Wang$^{1,*}$\\[0.6em]
\small $^{1}$School of Computer Engineering and Science, Shanghai University,\\
\small 99 Shangda Road, Shanghai 200444, People's Republic of China\\
\small $^{2}$Key Laboratory of Silicate Cultural Relics Conservation (Shanghai University),\\
\small Ministry of Education\\[0.4em]
\small E-mails: han\_yx@i.shu.edu.cn; 1649366224@qq.com; bingbignwang@shu.edu.cn\\
\small $^{*}$Corresponding authors: Yuexing Han and Bing Wang
}

\begin{document}
\begin{CJK}{UTF8}{gbsn}

\maketitle
\begin{center}
\small\itshape
This work has been submitted to the IEEE for possible publication.
Copyright may be transferred without notice, after which this version
may no longer be accessible.
\end{center}
\begin{abstract}
Generative models trained on a single image often struggle to balance global structural integrity and local diversity. Existing single-image generation methods commonly rely on random noise to drive the generation process and lack explicit global structural constraints, making the generated results prone to spatial structural misalignment when structural variations occur. To address the issue, Feature Rearrangement in Pre-Shape Space for Single-Image Generation (FRPSS) is proposed in this paper. The core of FRPSS is the Manifold Structural Rearrangement with Feature Augmentation on Geodesic Surface (MSR-FAGS) module. MSR-FAGS replaces the randomly initialized features of the low-scale generator with rearranged Pre-Shape features and uses the features to guide image generation at subsequent scales, thereby reducing the risk of structural misalignment. To support downstream tasks such as stylization, a Scale-adaptive Sliding-window Patch Extraction (SSPE) strategy is further designed, and a directional Contrastive Language-Image Pre-training supervision module with SSPE (CLIP-SSPE) is constructed. Qualitative and quantitative experiments demonstrate that FRPSS achieves the best Single Image Fréchet Inception Distance (SIFID) scores on all three datasets while maintaining competitive Learned Perceptual Image Patch Similarity (LPIPS). Further qualitative experiments verify the effectiveness of FRPSS across multiple downstream tasks with the CLIP-SSPE module.
\end{abstract}


\section{Introduction} \label{Sec:introduction}
Modern image generation models, particularly Generative Adversarial Networks (GANs) \cite{goodfellow2014generative, karras2020analyzing} and Denoising Diffusion Probabilistic Models (DDPMs) \cite{ho2020denoising, rombach2022high}, have demonstrated remarkable performance in synthesizing visual content with both fidelity and diversity. The generative models have been widely applied in image super-resolution, semantic scene synthesis, cross-modal content generation, and other fields \cite{azqadan2023predictive}. However, the strong generative capabilities of the models rely on massive amounts of training data, which greatly limits their applicability in specific domains. In fields such as material image analysis, historical artifact restoration, and private artistic creation, the acquisition and annotation of high-quality data are costly \cite{Kumar2024da, shorten2019survey}, which has become an important factor restricting the widespread application of generative models. Against the background, Single-Image Generation (SIG) has gradually developed into an important generation method for extremely data-scarce scenarios \cite{Shaham2019SinGAN}. SIG models can synthesize novel image variations that conform to the visual characteristics of the original image by capturing the internal statistics of a single input image. In addition, the single-image generation frameworks can support a series of downstream applications, such as text-guided style transfer, text-guided content generation, and paint-to-image translation. The downstream applications provide users with means for interactive and continuous semantic control over generated content and have certain practical value in customized graphic design, virtual reality, artistic creation, and other scenarios.

In recent years, research on SIG has gradually developed into two major categories. The first category is based on multi-scale GAN models \cite{Shaham2019SinGAN, Hinz_2021WACV_cosingan, NEURIPS2020_hpvaegan}, which employ a coarse-to-fine hierarchical framework to model local image patch distributions. Although the multi-scale methods can capture the distribution patterns of image patches, the generation process relies on random noise for initialization at low scales. Small structural deviations produced at low scales are continuously amplified during subsequent upsampling, and the progressive accumulation of errors can lead to structural misalignment in the generated results. The second category focuses on recently developed single-image diffusion models \cite{wang2025sindiffusion, he2026structdiff}. The diffusion-based methods train denoising diffusion models on a single image. Although the diffusion-based methods perform better than multi-scale models in image texture generation, the effective receptive field of the network usually needs to be restricted to prevent the diffusion model from overfitting to the single image. The limited spatial perception makes it difficult for the model to capture and maintain the global image structure, resulting in structural misalignment in the final generated results.

In addition, stylization \cite{kulikov2023sinddm, tumanyan2022splicing} is also an important downstream task of SIG frameworks. However, introducing cross-modal semantic supervision based on Contrastive Language-Image Pre-training (CLIP) \cite{patashnik2021styleclip, kwon2022clipstyler} into existing SIG frameworks still presents certain challenges. Images generated by models that rely on noise initialization are often unstable. Directly applying CLIP-based semantic constraints to the SIG models can not only aggravate error accumulation but also easily introduce hallucinated content, thereby making effective stylization difficult to achieve \cite{kulikov2023sinddm}.

To address the problems of existing single-image generation methods, FRPSS, a single-image generation method based on the Hierarchical Patch VAE-GAN framework \cite{NEURIPS2020_hpvaegan}, is proposed. The core component of FRPSS is the Manifold Structural Rearrangement with Feature Augmentation on Geodesic Surface (MSR-FAGS) module, which incorporates FAGS \cite{paskin2022kendall} to perform Geodesic surface interpolation in the Pre-Shape Space. The Pre-Shape Space theory \cite{kendall1984shape} is introduced into the single-image generation task to model features and exploit the structural prior information contained in a single training sample. The MSR-FAGS module not only provides subsequent generators with diverse and structurally plausible features, but also alleviates error accumulation during the multi-scale generation process.

For the stylization task in single-image generation frameworks, a Contrastive Language-Image Pre-training supervision module with Scale-adaptive Sliding-window Patch Extraction (CLIP-SSPE) is further designed. The CLIP-SSPE module adopts a scale-adaptive sliding-window patch extraction strategy and calculates the directional CLIP loss accordingly \cite{kwon2022clipstyler} to guide the image stylization process. The module guides the generation of stylized images corresponding to the text prompts provided by users.

\begin{figure*}[!htb] 
	\centering
	\includegraphics[width=\textwidth]{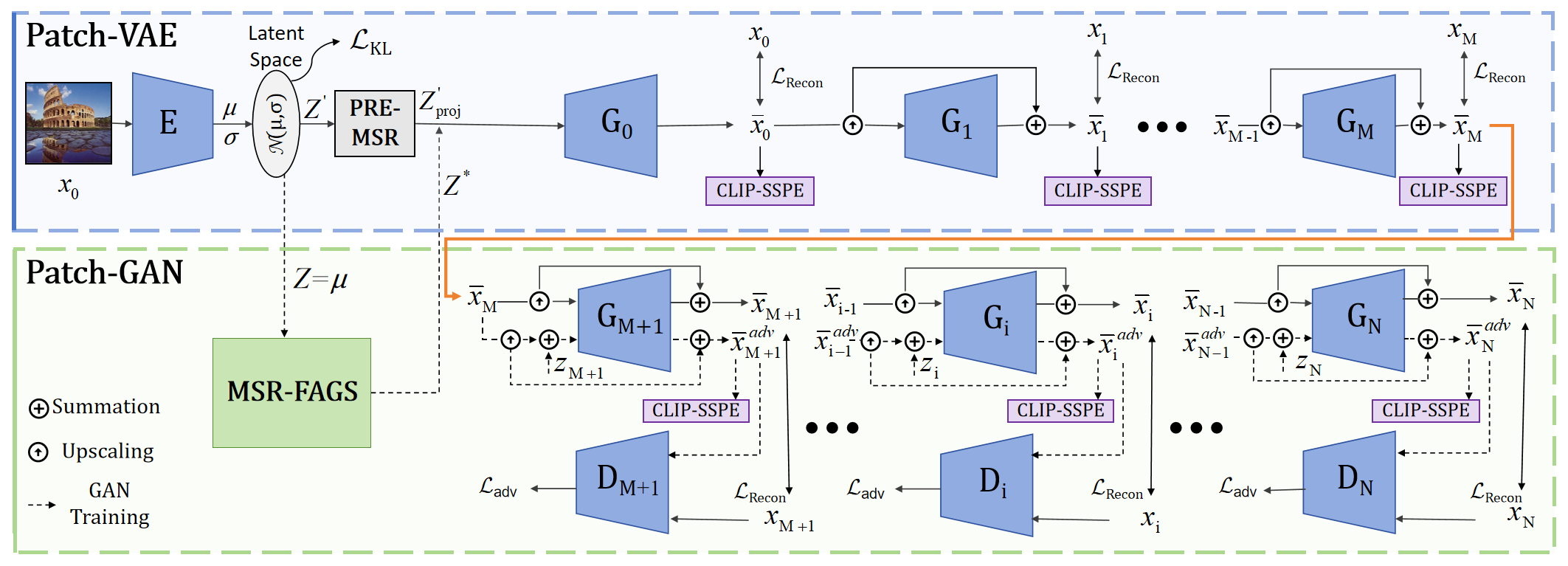} 
	\caption{Overall framework of FRPSS.}
	\label{fig:pvg} 
\end{figure*}
As shown in Fig. \ref{fig:pvg}, the MSR-FAGS module generates diverse and structurally plausible features, reduces the risk of structural misalignment in the generated results, and improves the fidelity of the generated images. The PRE-MSR module ensures consistency in representation form between its output features and the features produced by MSR-FAGS. The CLIP-SSPE module employs directional CLIP supervision for semantic guidance to achieve image stylization. Extensive qualitative and quantitative experiments demonstrate that FRPSS is highly competitive in the single-image generation task and can support multiple downstream tasks.

The main contributions of the paper are summarized as follows:
\begin{itemize}
    \item A single-image generation framework, FRPSS, is proposed. FRPSS can generate diverse images and support interactive control over the generation process.
    \item MSR-FAGS projects local features into the Pre-Shape Space and employs Geodesic surface interpolation to generate diverse and structurally plausible features.
    \item CLIP-SSPE constructs directional CLIP supervision during the multi-scale generation process and enables FRPSS to support downstream tasks.
\end{itemize}

\section{Related work}\label{Sec:related_work}
\subsection{Single-image generation}
Single-image generation methods synthesize diverse and realistic samples without additional training data by learning the statistics of local image patches from a single input image \cite{tirer2024deep}. Existing solutions in the field can be mainly divided into two categories: methods based on multi-scale Generative Adversarial Networks (GANs) \cite{zhang2021exsingan} and methods based on diffusion models \cite{nikankin2022sinfusion, kulikov2023sinddm, wang2025sindiffusion, he2026structdiff}.

SinGAN \cite{Shaham2019SinGAN} first introduced multi-scale GANs into the single-image generation task. In the architecture, cascaded generators progressively learn image structures and textures from coarse to fine scales. Subsequently, numerous variants based on the multi-scale GAN paradigm were proposed to further improve training efficiency, generation diversity, and other aspects of performance. For example, ConSinGAN \cite{Hinz_2021WACV_cosingan} accelerates the training process through a parallel multi-stage optimization strategy, whereas HP-VAE-GAN \cite{NEURIPS2020_hpvaegan} introduces a conditional Variational Autoencoder (VAE) branch to enhance the diversity of generated samples. To enhance the modeling capability for global structures, PetsGAN \cite{zhang2022petsgan} introduces external priors from a pre-trained generative model to enhance high-level semantic information in single-image generation. MOGAN \cite{chen2023mogan} models regions of interest and backgrounds through a morphology-aware mechanism to maintain reasonable image structures. SAMDSinGAN \cite{yildiz2024samdsingan} introduces a self-attention mechanism to capture long-range dependencies and global contextual information, thereby enhancing the modeling capability for overall image structures.

The second category is based on diffusion models for single-image generation \cite{ho2020denoising}. SinDDM \cite{kulikov2023sinddm} introduces a multi-scale cascaded architecture into single-image diffusion generation, whereas SinDiffusion \cite{wang2025sindiffusion} demonstrates that a diffusion network can effectively learn the internal data distribution at a single scale by using a local receptive field, thereby achieving high texture fidelity. Based on the single-image diffusion framework, the recent StructDiff \cite{he2026structdiff} introduces an adaptive receptive field mechanism to alleviate structural misalignment caused by a single receptive field. Recent studies have also begun to explore efficient training-free adaptation strategies \cite{qiu2026efficient}. In addition, recent studies have further extended single-image generation to more complex single-video generation tasks \cite{hou_ASC2025_union, verma2025novel}.

Although both multi-scale GAN methods and diffusion models have advanced the field of single-image generation, existing methods commonly adopt noise-based initialization strategies that inherently lack explicit structural priors. In multi-scale GAN architectures, the absence of reliable priors can lead to structural misalignment at the initial generation stage. The misalignment is progressively amplified when the generated results are upsampled to higher resolutions, leading to error accumulation. In diffusion models, the lack of explicit structural priors similarly limits the ability to control global spatial structures. In addition, diffusion models usually involve relatively high inference costs. To overcome the limitation caused by insufficient structural priors, FRPSS replaces random noise initialization with rearranged latent features in the Pre-Shape Space at the low-scale generation stage, thereby maintaining the stability of structural layouts in subsequent generated images.

\subsection{Single-image stylized generation}
Many existing stylization methods employ large-scale vision-language models \cite{radford2021learning} to guide image editing and generation. In early studies, StyleCLIP \cite{patashnik2021styleclip} employs a CLIP-based loss that minimizes the cosine distance between the embeddings of a generated image and a target text prompt to guide text-driven image manipulation. However, directly applying a global CLIP loss may lead to unstable optimization and degraded image quality. To address the limitation, StyleGAN-NADA \cite{gal2022stylegan_nada} introduces a directional CLIP loss. The loss constrains the semantic change direction of an image to be consistent with the semantic change direction of text. Subsequent studies further extend directional CLIP guidance to text-guided image stylization and manipulation. For example, CLIPStyler \cite{kwon2022clipstyler} proposes a patch-wise CLIP loss to guide local texture stylization. Similarly, DiffusionCLIP \cite{kim2022diffusionclip} employs a directional CLIP loss for stylized generation in diffusion models.

To enable interactive control over single-image generation models, many methods incorporate CLIP-based supervision into the generation process. For example, SinDDM \cite{kulikov2023sinddm} uses the gradients of a global CLIP loss to update the predicted clean image during diffusion sampling, enabling text-guided image stylization. However, SinDDM simultaneously adjusts image structure and texture during iterative denoising. The coupling makes semantic supervision prone to disturbing the spatial layout of the image, which may lead to hallucinated content and structural distortion. To alleviate the problem, FRPSS applies a directional CLIP loss \cite{kwon2022clipstyler} to the generated result at each scale. Based on a stable image structure, FRPSS uses the directional CLIP loss to guide image generation and reduce interference with the original spatial layout during semantic injection, thereby reducing artifacts and hallucinated content in the generated results.

\subsection{The Shape Space theory}
The Shape Space theory was introduced by Kendall \cite{kendall1984shape}. Under the theory, shape is defined as geometric information that remains invariant to translation, scaling, and rotation. Compared with the Shape Space, the Pre-Shape Space provides a simplified representation that removes only translation and scaling while retaining rotational information. The Pre-Shape Space also supports corresponding feature augmentation methods. Therefore, FRPSS adopts the Pre-Shape Space to model features. Given a feature vector $U = \{u_1, \cdots, u_m\} \in \mathbb{R}^m$, landmarks are constructed through a simple lifting operation by setting $v_i = u_i$. The resulting landmark matrix is $A = [(u_1, v_1), \cdots, (u_m, v_m)] \in \mathbb{R}^{2 \times m}$. The centered form $A'$ is then computed and normalized \cite{kendall1984shape} to obtain the Pre-Shape $\tau$:
\begin{equation}\label{eq:preshape}
\begin{aligned}
A' &= [(u_1-\bar{u}, v_1-\bar{v}), \cdots, (u_m-\bar{u}, v_m-\bar{v})], \\
\tau &= \frac{A'}{\|A'\|} \in \mathbb{S}^{2m-3},
\end{aligned}
\end{equation}
where $\bar{u}$ and $\bar{v}$ denote the means of the landmarks along the two dimensions, respectively, and $\|\cdot\|$ denotes the Euclidean norm. All Pre-Shapes $\tau$ lie on the hypersphere $\mathbb{S}^{2m-3}$ corresponding to the Pre-Shape Space.

Following the interpolation method of Han et al. \cite{han2010recognition}, FRPSS adopts Geodesic interpolation in the Pre-Shape Space to augment the Pre-Shape samples on the manifold. For two Pre-Shapes $\tau_a$ and $\tau_b$, let
$\theta=\arccos(\langle\tau_a,\tau_b\rangle)$
denote the Geodesic distance between them in the Pre-Shape Space. The interpolation point on the shortest Geodesic curve between the two Pre-Shapes is given by $\mathbb{G}_{slerp}$ \cite{han2010recognition, han2014recognizing}:

\begin{equation}\label{eq:slerp}
	\mathbb{G}_{slerp}(\tau_a,\tau_b,t)
	=
	\cos(t\theta)\tau_a
	+
	\sin(t\theta)
	\frac{\tau_b-\tau_a\cos\theta}{\sin\theta},
\end{equation}

where $t\in[0,1]$ denotes the normalized interpolation position along the shortest Geodesic curve. When $t=0$, the interpolation result is $\tau_a$, and when $t=1$, the interpolation result is $\tau_b$.

To perform interpolation on a Geodesic surface constructed from a set of Pre-Shapes
$\mathcal{T} = \{\tau_1, \dots, \tau_K\}$,
FRPSS employs the FAGS algorithm \cite{paskin2022kendall} for iterative Geodesic interpolation.
Given a set of weights
$\boldsymbol{\omega} = \{\omega_1, \dots, \omega_K\}$
sampled from a Dirichlet distribution and satisfying
$\sum_{i=1}^{K} \omega_i = 1$,
FAGS synthesizes a Pre-Shape $\tau_{syn}$ through iterative Geodesic interpolation.
The initialization is set as $\tilde{\tau}_1 = \tau_1$, and the subsequent iterations are defined as follows \cite{paskin2022kendall}:

\begin{equation}\label{eq:fags}
	\tilde{\tau}_j =
	\mathbb{G}_{slerp}\left(
	\tilde{\tau}_{j-1},
	\tau_j,
	\frac{\omega_j}{\sum_{k=1}^{j} \omega_k}
	\right),
	\quad \text{for } j=2, \dots, K.
\end{equation}

The final output $\tau_{syn} = \tilde{\tau}_K$ represents a first-order approximation to the weighted Fr\'echet mean of the set of Pre-Shapes under the weights $\boldsymbol{\omega}$.

\section{Methodology} \label{Sec:method}
Fig. \ref{fig:pvg} illustrates the two stages of the FRPSS framework: the Patch-VAE stage and the Patch-GAN stage. The Patch-VAE stage consists of an encoder $E$, the PRE-MSR module, and generators $G_0, \cdots, G_M$. The Patch-VAE stage is designed to reconstruct images at low scales. PRE-MSR ensures that the feature vectors extracted by $E$ have the same representation form as the output features of MSR-FAGS. The Patch-GAN stage consists of the MSR-FAGS module, generators $G_{M+1}, \cdots, G_i, \cdots, G_N$, and the corresponding discriminators $D_{M+1}, \cdots, D_i, \cdots, D_N$. The Patch-GAN stage is designed to generate diverse images with high-frequency details. MSR-FAGS performs feature augmentation by taking the mean feature $Z$ produced by $E$ as input and generating the augmented feature $Z^*$. In addition, FRPSS introduces cross-modal semantic supervision into the generation process. The CLIP-SSPE module calculates the directional CLIP loss to guide image generation according to user-provided text prompts.

\subsection{Patch-VAE stage}
Given a single input image $x$, FRPSS first resizes the image to the predefined largest scale $x_N$ using bilinear interpolation. The resulting image $x_N$ is then progressively downsampled to obtain a set of images at different scales, denoted as $\{x_i\}_{i=0}^N$. Here, $x_i$ denotes the downsampled image at scale $i$, and $x_0$ corresponds to the smallest scale. Each image $x_i$ is associated with a generator $G_i$ for image generation at the corresponding scale. The downsampling ratio between two adjacent scales is set to 0.75.

The encoder $E$ first extracts the features of the input image $x_0$ through a convolutional feature extraction module. Two convolutional branches then output the mean feature $\mu \in \mathbb{R}^{C \times H \times W}$ and the standard deviation feature $\sigma \in \mathbb{R}^{C \times H \times W}$, respectively. Here, $C$ denotes the channel dimension, while $H$ and $W$ denote the height and width of the feature space, respectively. For any spatial position $j \in \{1, \dots, H \times W\}$ in the feature space, the $C$-dimensional vectors $\mu_j$ and $\sigma_j$ jointly define the Gaussian distribution of the image patch corresponding to the position. Following the setting of HP-VAE-GAN \cite{NEURIPS2020_hpvaegan}, the patch-level Kullback--Leibler ($KL$) divergence can be independently computed at position $j$. By summing the divergence values over all local positions in the feature map, the total $KL$ loss for training the Patch-VAE stage is obtained as follows:
\begin{equation}\label{e1}
	\mathcal{L}_{KL}(x) = \sum_{j=1}^{H \times W}KL[\mathcal{N}(\mu_j, \sigma_j^2) \| \mathcal{N}(0,I)].
\end{equation}
Then, standard reparameterization sampling is performed on $\mu$ and $\sigma$ output by $E$ to obtain $Z'\in \mathbb{R}^{C\times H\times W}$:
\begin{equation} \label{e2}
    Z' = \mu + \sigma \odot \epsilon, 
\end{equation}
where $\epsilon \sim \mathcal{N}(0, I)$, and $\odot$ denotes element-wise multiplication.

\begin{figure*}[!htb] 
	\centering
	\includegraphics[width=\textwidth]{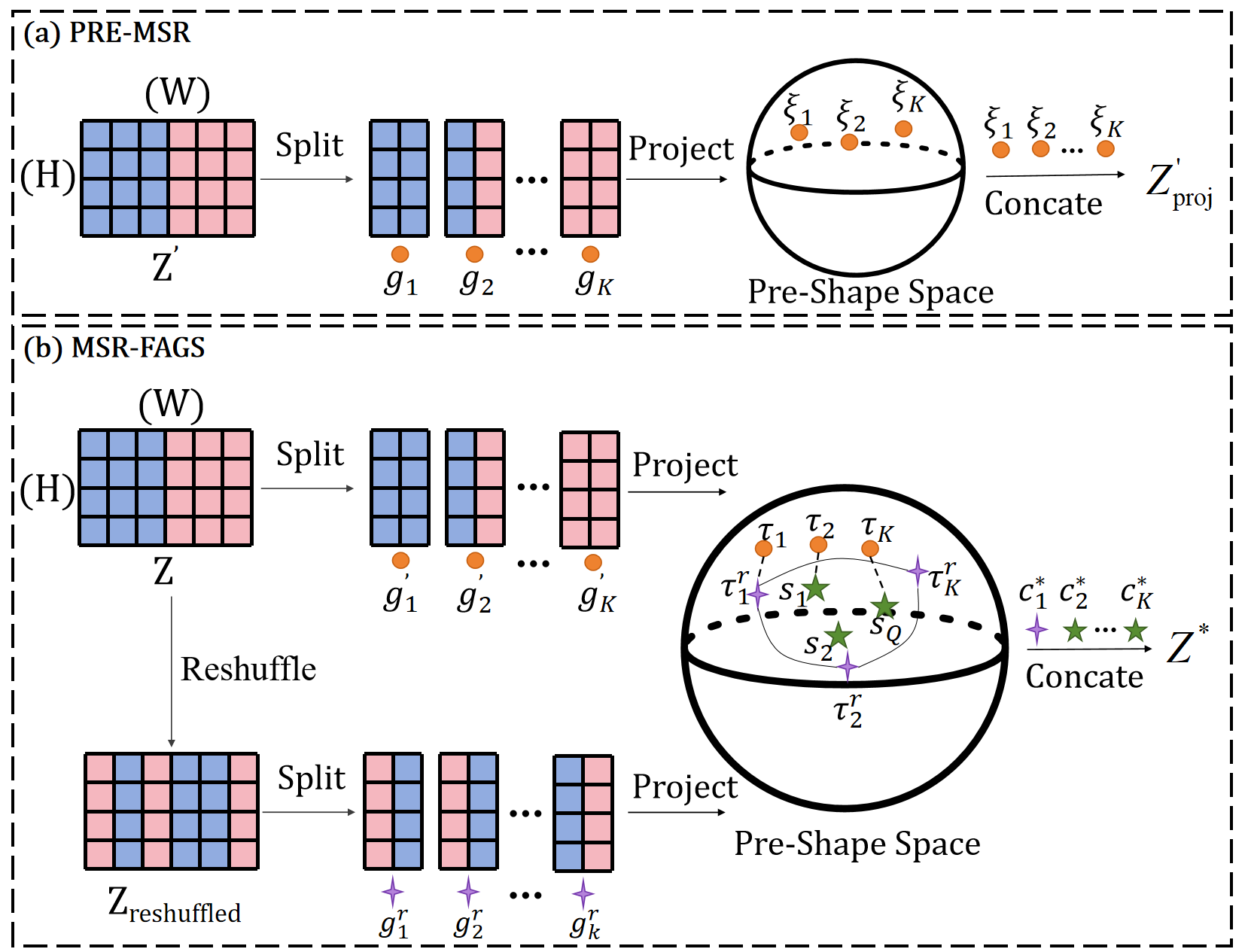} 
	\caption{Feature projection and rearrangement based on the Pre-Shape Space. (a) PRE-MSR module and (b) MSR-FAGS module.}
	\label{fig:msr}
\end{figure*}

To make $Z'$ have the same representation form as the output feature $Z^*$ of MSR-FAGS, the PRE-MSR module divides $Z'$ along the width $W$ into $K$ consecutive features of width $\alpha$. The features are denoted as $\mathcal{G}=(g_1, \cdots,g_k ,\cdots, g_K)$, where $K=W/\alpha$ and $g_k \in \mathbb{R}^{C \times H \times \alpha}$. Subsequently, according to Eq. \eqref{eq:preshape}, each $g_k$ is projected into the Pre-Shape Space to obtain $\xi_k \in \mathbb{R}^{2C \times H \times \alpha}$. The projected features are then concatenated in order to obtain $Z^{'}_{proj}=(\xi_1, \cdots, \xi_k, \cdots, \xi_K) \in \mathbb{R}^{2C \times H \times W}$. The entire process is shown in Fig. \ref{fig:msr}(a). The detailed process of the PRE-MSR module is shown in Algorithm \ref{alg:a1}.
\begin{algorithm}[!htb]
	\caption{PRE-MSR Module in the Patch-VAE training stage.}
	\label{alg:a1}
	\SetKwInOut{Input}{Input}
	\SetKwInOut{Output}{Output}
	\SetKwFunction{Project}{Project}
	\SetKwFunction{Concat}{Concat}
	\SetKwFunction{Reshape}{Reshape}
	\Input{Latent Feature $Z' \in \mathbb{R}^{C \times H \times W}$, Group size $\alpha$}
	\Output{Projected Feature $Z^{'}_{proj} \in \mathbb{R}^{2C \times H \times W}$}
	\tcp{1. Spatial Split}
	Split $Z'$ along width into spatial groups of size $\alpha$\;
	Let $\mathcal{G} = \{g_1, g_2, \dots, g_K\}$ be the sequence of groups, where $K = W/\alpha$\;
	Initialize projected list $\mathcal{P} \leftarrow []$\;
	\tcp{2. Project to the Pre-Shape Space}
	\For{$k \leftarrow 1$ \KwTo $K$}{
		\tcp{Project $g_k \in \mathbb{R}^{C \times H \times \alpha}$ to a manifold point}
		$\xi_k \leftarrow \Project(g_k)$\;
		Append $\xi_k$ to $\mathcal{P}$\;
	}
	\tcp{3. Assembly \& Feature Lifting}
	$Z_{flat} \leftarrow \Concat(\mathcal{P})$\;
	$Z^{'}_{proj} \leftarrow \Reshape(Z_{flat})$\;
	\Return $Z^{'}_{proj}$\;
\end{algorithm}

$Z^{'}_{proj}$ is input into the generator $G_0$ to obtain the image $\bar{x}_0$, where $\bar{x}_0 = G_0(Z^{'}_{proj})$. To progressively reconstruct the original image from low scales, details are gradually added at the subsequent intermediate scales $i \in \{1, \cdots, M\}$. The generators $G_0, G_1, \cdots, G_M$ correspond to different generation scales. $G_0$ reconstructs the low-scale image according to the latent feature output by the encoder $E$. The generators $G_1, \cdots, G_M$ predict the residual of the current scale relative to the generated image at the previous scale. During training at scale $i$, the encoder $E$, the generator $G_0$, and $G_i$ are jointly optimized, while the parameters of the intermediate generators $G_1, \dots, G_{i-1}$ remain frozen. Since the encoder $E$ is continuously updated during training at different scales, its output latent feature also changes accordingly. Therefore, $G_0$ needs to be jointly optimized with $E$ to maintain the mapping relationship between the latent representation and the reconstruction result at the lowest scale. The intermediate generators that have already been trained remain frozen to prevent subsequent scale training from changing their learned scale-specific residual mappings. Let $\bar{x}_{i-1}$ denote the reconstruction result at the previous scale. The previous reconstruction $\bar{x}_{i-1}$ is first upsampled using bilinear interpolation, denoted by $\uparrow$, to match the image resolution at scale $i$. The upsampled result is then input into the generator $G_i$. $\bar{x}_i$ can be expressed as the sum of the upsampled result and the residual predicted by the generator \cite{NEURIPS2020_hpvaegan}:
\begin{equation}\label{e3}
\bar{x}_i = {\uparrow}\bar{x}_{i-1} + G_i({\uparrow}\bar{x}_{i-1}).
\end{equation}
The reconstruction loss between the generated image $\bar{x}_i$ and the real image $x_i$ is used to calculate the difference between them:
\begin{equation}\label{e4}
\mathcal{L}_{Recon}(\bar{x}_i, x_i) = \|\bar{x}_i - x_i\|_2,
\end{equation}
where $\|\cdot\|_2$ denotes the $L_2$ norm.

Following the $\beta$-VAE framework proposed by Higgins et al. \cite{higgins2017betavae}, the hyperparameter $\beta$ is introduced in the Patch-VAE stage to weight the $KL$ divergence loss. Finally, the total loss of the Patch-VAE stage is defined as follows \cite{NEURIPS2020_hpvaegan}:
\begin{equation}\label{e5}
\mathcal{L}_{VAE}(x_0, \bar{x}_i, x_i) = \lambda_{r}\mathcal{L}_{Recon}(\bar{x}_i, x_i) + \beta \mathcal{L}_{KL}(x_0), i\in \{0,\cdots, M\}.
\end{equation}
where $\lambda_{r}$ is a hyperparameter used to balance the contributions of the losses.

\subsection{Patch-GAN stage}\label{sec:patchGan}
When the scale $i \ge M+1$, the model enters the Patch-GAN stage to generate diverse images with high-frequency texture information. At the stage, the parameters of the encoder $E$ and the generators $G_0, \cdots, G_M$ trained in the Patch-VAE stage are frozen.

The mean feature $\mu$ output by the encoder $E$ serves as the input vector $Z$ of the MSR-FAGS module, denoted as $Z=\mu$. Here, only the mean feature is used, and reparameterization sampling is no longer performed, because $\mu$ encodes the structural prior of the original image. The deterministic representation avoids the random noise perturbation introduced by the reparameterization sampling process and provides a stable reference for the subsequent projection into the Pre-Shape Space. The MSR-FAGS module rearranges the feature $Z$ in the Pre-Shape Space to generate a new feature $Z^*$ with structural variations. Subsequently, $Z^*$ is input into the generators $G_0, \cdots, G_M$ trained in the Patch-VAE stage to obtain the image $\bar{x}_M$ at scale $M$. Based on the result, scale-by-scale adversarial training is performed at the subsequent scales $M+1, \cdots, N$.

As shown in the Patch-GAN stage of Fig. \ref{fig:pvg}, for scale $i$, the generated image $\bar{x}_{i-1}^{adv}$ at the previous scale is first upsampled to the current scale. Next, the upsampled result is added to the noise $z_i$ at the current scale, and the sum is input into the generator $G_i$ to predict the high-frequency texture residual. Finally, the residual output by $G_i$ is added to the upsampled $\bar{x}_{i-1}^{adv}$ to obtain the generated image $\bar{x}_i^{adv}$ at the current scale $i$. Therefore, the generation process of $\bar{x}_i^{adv}$ can be defined as follows:
\begin{equation}\label{e6} 
\bar{x}_i^{adv} = {\uparrow}\bar{x}_{i-1}^{adv} + G_i({\uparrow}\bar{x}_{i-1}^{adv} + z_i).
\end{equation}
The WGAN-GP loss \cite{gulrajani2017wgangp} is adopted for adversarial training at the stage:
\begin{equation}\label{e7}
	\mathcal{L}_{adv}(z_i, x_i) = \min_{G_i} \max_{D_i} \left\{ \mathbb{E}[D_i(x_i)] - \mathbb{E}[D_i(\bar{x}_i^{adv})] - \lambda \mathbb{E}\left[ (\|\nabla_{\hat{x}} D_i(\hat{x})\|_2 - 1)^2 \right] \right\},
\end{equation}
where $D_i$ denotes the discriminator at the current scale $i$, $\nabla$ denotes the gradient operator, and $\lambda$ is the gradient penalty coefficient. Here, $x_i$ denotes the real image at the current scale $i$, while $\hat{x}$ denotes the uniformly interpolated result between the real sample $x_i$ and the generated sample $\bar{x}_i^{adv}$. The calculation of $\hat{x}$ can be expressed as follows \cite{gulrajani2017wgangp}:
\begin{equation}\label{e8} 
	\hat{x} = \epsilon x_i + (1 - \epsilon) \bar{x}_i^{adv}, \quad \epsilon \sim U(0, 1).
\end{equation}

In addition, the reconstruction loss used in the Patch-VAE stage is also adopted in the Patch-GAN stage. Therefore, the total adversarial training loss is as follows:
\begin{equation}\label{e9}
\mathcal{L}_{GAN}(z_i, \bar{x}_i, x_i) = \lambda_{r}\mathcal{L}_{Recon}(\bar{x}_i, x_i) + \beta_{adv}\mathcal{L}_{adv}(z_i, x_i), i \in \{M+1, \cdots, N\},
\end{equation}
where $\lambda_{r}$ and $\beta_{adv}$ are hyperparameters used to balance the contributions of the losses. The generation process of $\bar{x}_i$ is consistent with that in the Patch-VAE stage.

\subsection{Manifold Structural Rearrangement with FAGS}
The MSR-FAGS module generates diverse and structurally plausible features. Algorithm \ref{alg:a2} describes the complete execution process of the MSR-FAGS module.
\begin{algorithm}[!htb]
	\caption{MSR-FAGS module in the Patch-GAN stage.}
	\label{alg:a2}
	\SetKwInOut{Input}{Input}
	\SetKwInOut{Output}{Output}
	\SetKwFunction{Project}{Project}
	\SetKwFunction{FAGS}{FAGS}
	\SetKwFunction{Shuffle}{GlobalPermute}
	\SetKwFunction{Concat}{Concat}
	
	\Input{Feature tensor $Z \in \mathbb{R}^{C \times H \times W}$, Group size $\alpha$, Number of interpolated features $Q$}
	\Output{Enhanced Feature $Z^* \in \mathbb{R}^{2C \times H \times W}$}
	
	\tcp{1. Spatial Decomposition \& Anchor Path Construction}
	Split $Z$ along width $W$ into slices and group them by size $\alpha$\;
	Let $\mathcal{G}' = \{g'_1, \dots, g'_k, \dots, g'_K\}$ be the ordered groups, where $K = W/\alpha$\;
	\For{$k \leftarrow 1$ \KwTo $K$}{
		$\tau_k \leftarrow \Project(g'_k)$ \tcp*{Map to the Pre-Shape Space}
	}
	Set $\mathcal{T} \leftarrow \{\tau_1, \dots, \tau_K\}$\;
	
	\tcp{2. Candidate Path via Global Shuffling}
	$Z_{reshuffled} \leftarrow \Shuffle(Z, \text{dim}=W)$ \tcp*{Global random reshuffling}
	Split $Z_{reshuffled}$ into $\mathcal{G}^r = \{g^{r}_1, \dots, g^{r}_k, \dots, g^{r}_K\}$\;
	\For{$k \leftarrow 1$ \KwTo $K$}{
		$\tau^{r}_k \leftarrow \Project(g^{r}_k)$ \tcp*{Map to the Pre-Shape Space}
	}
	Set Candidate Base $\mathcal{T}^{r} \leftarrow \{\tau^{r}_1, \dots, \tau^{r}_k, \dots, \tau^{r}_K\}$\;
	
	\tcp{3. Manifold Expansion via FAGS}
	Initialize $\mathcal{S} \leftarrow \emptyset$\;
	\For{$j \leftarrow 1$ \KwTo $Q$}{
		Sample weights $\boldsymbol{\omega} \sim \text{Dir}(\mathbf{1}_K)$\;
		$s_j \leftarrow \FAGS(\mathcal{T}^{r}, \boldsymbol{\omega})$ \tcp*{Iterative Geodesic Interpolation}
		Add $s_j$ to $\mathcal{S}$\;
	}
	Set $\mathcal{C} \leftarrow \Concat(\mathcal{T}^{r}, \mathcal{S})$\;
	
	\tcp{4. Feature Reconstruction via Shortest Geodesic Matching}
	Initialize reconstructed list $\mathcal{L} \leftarrow []$\;
	\For{$k \leftarrow 1$ \KwTo $K$}{
		\tcp{Search for best feature match in $\mathcal{C}$}
		$c^*_{k} \leftarrow \underset{c \in \mathcal{C}}{\arg\min} \ \arccos\left( \langle \tau_k, c \rangle \right)$\;
		Append $c^*_{k}$ to $\mathcal{L}$\;
	}
	$Z^* \leftarrow \Concat(\mathcal{L})$ \tcp*{Reshape and Assembly (2C channels)}
	\Return $Z^*$\;
\end{algorithm}

To overcome the data limitation of a single sample, MSR-FAGS adopts the feature decomposition strategy of MultiOSG \cite{gou2025multiosg}. As shown in Fig. \ref{fig:msr}(b), the feature $Z$ is uniformly divided along the width $W$ into $K$ consecutive features in the same manner as the PRE-MSR module. The resulting features form an ordered set $\mathcal{G}' = \{g'_1, \dots, g'_k, \dots, g'_K\}$, where $g'_k \in \mathbb{R}^{C \times H \times \alpha}$. Here, $K = W / \alpha$ denotes the number of features in $\mathcal{G}'$, and $\alpha$ denotes the width of each feature. Subsequently, each feature $g'_k$ is projected into the Pre-Shape Space in order according to Eq. \eqref{eq:preshape}. The projection yields the feature set $\mathcal{T} = \{\tau_1, \dots, \tau_k, \dots, \tau_K\}$, where $\tau_k \in \mathbb{R}^{2C \times H \times \alpha}$.

To introduce feature diversity, MSR-FAGS first splits the feature $Z$ along the width $W$ into $W$ column features of size $C \times H$. The order of the $W$ features is then randomly shuffled, and the features are concatenated again to obtain $Z_{reshuffled}$. Subsequently, MSR-FAGS divides $Z_{reshuffled}$ along the width $W$ into $K$ consecutive features to obtain the set $\mathcal{G}^{r} = \{g^{r}_1, \dots, g^{r}_k, \dots, g^{r}_K\}$, where $g^{r}_k \in \mathbb{R}^{C \times H \times \alpha}$. Each feature in $\mathcal{G}^r$ is also projected into the Pre-Shape Space according to Eq. \eqref{eq:preshape}. The projection yields the feature set $\mathcal{T}^r = \{\tau^{r}_1, \dots, \tau^{r}_k, \dots, \tau^{r}_K\}$, where $\tau^{r}_k \in \mathbb{R}^{2C \times H \times \alpha}$.

To expand the set $\mathcal{T}^{r}$, iterative interpolation is performed on the Geodesic surface constructed by the Pre-Shapes in $\mathcal{T}^{r}$ according to Eq. \eqref{eq:fags}. In each new feature generation step, MSR-FAGS first samples a set of weights $\boldsymbol{\omega}$ from the Dirichlet distribution to determine the contribution proportion of each feature in $\mathcal{T}^{r}$. Using the sampled weights $\boldsymbol{\omega}$, iterative Geodesic interpolation is then performed on the features in $\mathcal{T}^{r}$ on the Geodesic surface. The interpolation generates a new feature $s_j$. After repeating the above process $Q$ times, the module obtains a new feature set $\mathcal{S}=\{s_1, \cdots , s_j, \cdots, s_Q\}$ and merges it with $\mathcal{T}^{r}$ to obtain the expanded feature set $\mathcal{C}$.

MSR-FAGS then matches the features in $\mathcal{T}$ with those in $\mathcal{C}$ using the shortest Geodesic distance. For each feature $\tau_k$ corresponding to a position in $\mathcal{T}$, MSR-FAGS searches $\mathcal{C}$ for the feature $c^*_{k}$ with the shortest Geodesic distance to $\tau_k$. The matching allows the recombined feature to preserve the spatial coherence of the original image as much as possible while introducing variations. Here, MSR-FAGS adopts the shortest Geodesic distance in the Pre-Shape Space \cite{han2010recognition} as the matching criterion. The corresponding matching process is defined as follows:
\begin{equation}\label{e10}
	c^*_{k} = \underset{c \in \mathcal{C}}{\arg\min} \ \arccos\left( \langle \tau_k, c \rangle\right).
\end{equation}
MSR-FAGS concatenates the matched $K$ features in order along the width $W$ to obtain $Z^*=(c^*_1, \dots, c^*_k, \dots, c^*_K) \in \mathbb{R}^{2C \times H \times W}$ and inputs $Z^*$ into $G_0$. Finally, the generators are trained according to the process in Section \ref{sec:patchGan}.

\subsection{Stylized generation}
\begin{figure*}[!htb] 
	\centering
	\includegraphics[width=\textwidth]{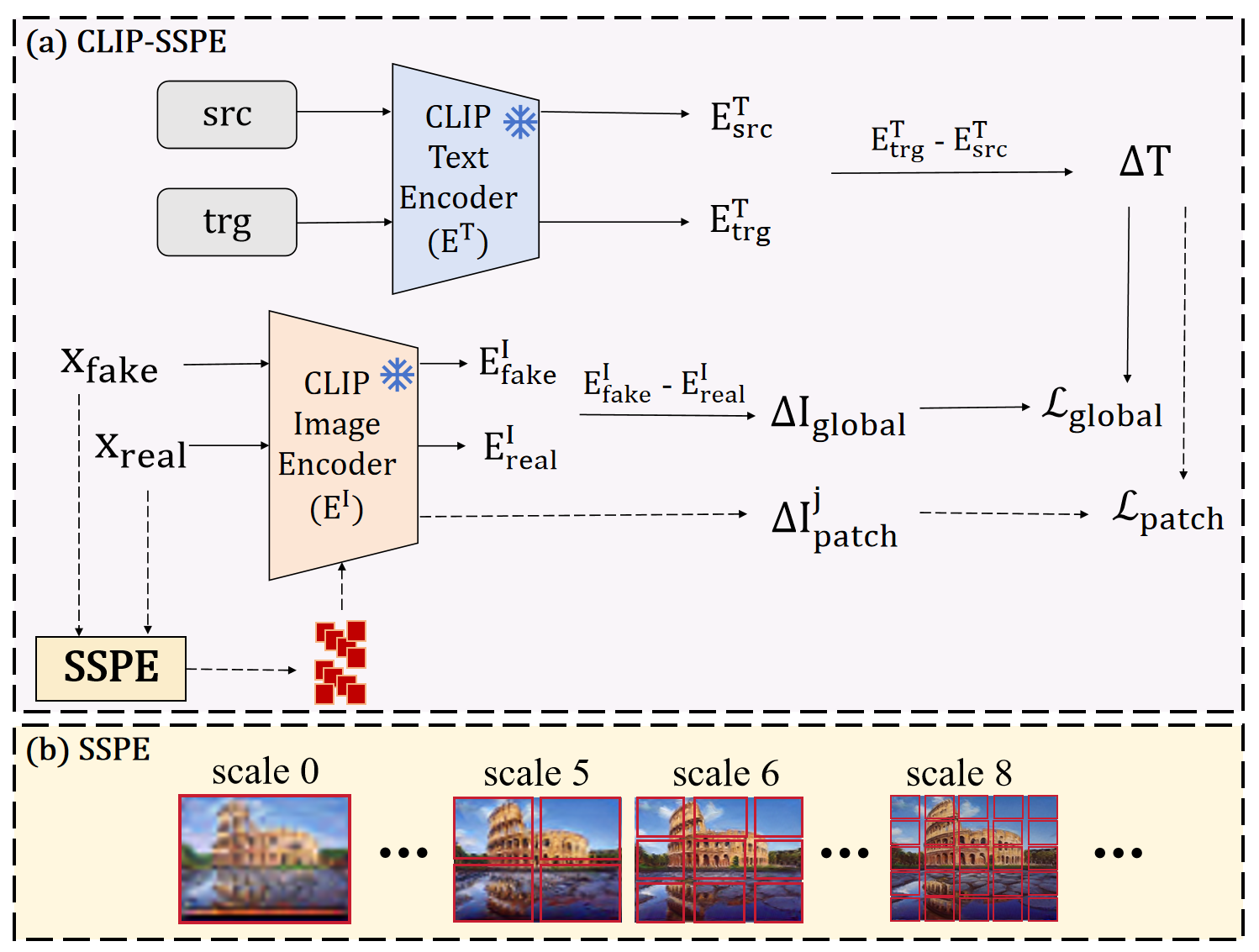} 
	\caption{Illustration of the CLIP stylization loss calculation based on the CLIP-SSPE module.}
	\label{fig:clip} 
\end{figure*}

To further explore the applicability of FRPSS in practical scenarios, FRPSS is extended to downstream stylization tasks. In the stylization task, FRPSS uses a pre-trained CLIP model to provide cross-modal semantic priors. As shown in Fig. \ref{fig:pvg}, the CLIP-SSPE module is incorporated into each generator $G_i$ in both the Patch-VAE and Patch-GAN stages. FRPSS progressively accumulates the generation results at different scales to form the final generated image rather than relying on the last generator to independently generate the final image. As a result, semantic constraints applied only at the final scale cannot sufficiently affect the visual features already formed at previous scales. CLIP-SSPE therefore progressively applies semantic supervision at each scale and introduces the target semantics scale by scale during the multi-scale generation process. As shown in Fig. \ref{fig:clip}, the CLIP-SSPE module contains a frozen CLIP text encoder $E^T$, a frozen image encoder $E^I$, and the SSPE module. $E^T$ and $E^I$ encode the text and image, respectively, and project them into the joint embedding space. The SSPE module adaptively extracts image patches according to the generated images at different scales. Following the setting of CLIPStyler \cite{kwon2022clipstyler}, CLIP-SSPE adopts a directional CLIP supervision mechanism. The module constructs the directional constraints $\mathcal{L}_{global}$ and $\mathcal{L}_{patch}$ at the global and patch-wise levels, respectively. The directional constraints make the visual change direction from the real image to the generated image consistent with the semantic change direction from the source text prompt to the target text prompt.

The inputs of CLIP-SSPE include the real image $x_{real}$, the generated image $x_{fake}$ at the current scale, and the source text prompt $src$ and target text prompt $trg$ provided by the user. Here, $src$ indicates that the input image is an image, while $trg$ indicates the target style to be changed. CLIP-SSPE inputs the two text prompts into the frozen CLIP text encoder $E^T$ to obtain the corresponding text features $E^{T}_{src}$ and $E^{T}_{trg}$. To align the image transformation direction with the text semantic transformation direction in the CLIP joint embedding space, the text direction vector $\Delta T$ is defined based on CLIPStyler \cite{kwon2022clipstyler} as follows:
\begin{equation}\label{e11}
\Delta T = E^{T}_{trg} - E^{T}_{src}.
\end{equation}

At any scale, the CLIP-SSPE module uses the complete generated image and the corresponding real image to calculate the global directional CLIP loss. The frozen image encoder $E^I$ extracts the features $E^{I}_{fake}$ and $E^{I}_{real}$ from $x_{fake}$ and $x_{real}$, respectively. Following CLIPStyler \cite{kwon2022clipstyler}, the global image direction vector $\Delta I_{global}$ is defined as follows:
\begin{equation}\label{e12}
\Delta I_{global} = E^{I}_{fake} - E^{I}_{real}.
\end{equation}
The global directional loss $\mathcal{L}_{global}$ is used to maximize the cosine similarity between $\Delta I_{global}$ and $\Delta T$:
\begin{equation}\label{e13}
\mathcal{L}_{global} = 1 - \frac{\Delta I_{global} \cdot \Delta T}{\|\Delta I_{global}\| \cdot \|\Delta T\|}.
\end{equation}

For the calculation of the patch-wise directional CLIP constraint, SSPE extracts different numbers of image patches at different scales. Assume that $H \times W$ denotes the image resolution at the current scale. To perform local image patch extraction, SSPE adopts a sliding crop mechanism. The size $r$ of the sliding crop window is adaptively adjusted according to the current image resolution and is restricted to the predefined range $[64, 128]$. The size $r$ of the sliding crop window is calculated as follows:
\begin{equation}\label{e14}
	r = \max\left(64, \min\left(\left\lfloor \frac{\min(H, W)}{2} \right\rfloor, 128\right)\right).
\end{equation}
To ensure a sufficient overlap ratio between the cropped image patches for maintaining texture continuity, the sliding stride $s$ is defined as half of the window size:
\begin{equation}\label{e15}
	s = \frac{r}{2}.
\end{equation}
Based on the above configuration, for an image with a resolution of $H \times W$, the total number $D$ of image patches extracted by the sliding window is:
\begin{equation}\label{e16}
	D = \left\lfloor \frac{H}{s} \right\rfloor  \times  \left\lfloor \frac{W}{s} \right\rfloor.
\end{equation}
It is worth noting that when the image size is smaller than the minimum window limit or when the image patch extraction condition cannot be satisfied, SSPE no longer performs local cropping. Instead, SSPE directly uses the complete generated image and real image to calculate the directional constraint. In the case, the patch-wise constraint degenerates into the global constraint.

During image patch cropping, let $p_{fake}^{j}$ and $p_{real}^{j}$ denote the $j$-th corresponding image patches cropped from the generated image and the real image, respectively. The image direction vector $\Delta I_{patch}^{j}$ between the two patches is defined as follows \cite{kwon2022clipstyler}:
\begin{equation}\label{e17}
\Delta I_{patch}^j = E^I(p_{fake}^j) - E^I(p_{real}^j).
\end{equation}
Subsequently, CLIP-SSPE calculates the cosine distance between $\Delta I_{patch}^j$ and the text direction vector $\Delta T$. The arithmetic mean of the cosine losses corresponding to the $D$ image patches cropped at the current scale is then taken to obtain the final patch-wise directional loss $\mathcal{L}_{patch}$ \cite{kwon2022clipstyler}:
\begin{equation}\label{e18}
	\mathcal{L}_{patch} = \frac{1}{D} \sum_{j=1}^{D} \left( 1 - \frac{\Delta I_{patch}^j \cdot \Delta T}{\|\Delta I_{patch}^j\| \cdot \|\Delta T\|} \right).
\end{equation}

Finally, the Total Variation (TV) regularization loss $\mathcal{L}_{tv}$ \cite{kwon2022clipstyler} is also introduced into the CLIP-SSPE module. The overall CLIP stylization loss is:
\begin{equation}\label{e19}
\mathcal{L}_{clip} = \lambda_{global}\mathcal{L}_{global} + \lambda_{patch}\mathcal{L}_{patch} + \lambda_{tv}\mathcal{L}_{tv}.
\end{equation}
where $\lambda_{global}$, $\lambda_{patch}$, and $\lambda_{tv}$ are hyperparameters used to balance the contributions of the losses. In the experiments, $\lambda_{global}$, $\lambda_{patch}$, and $\lambda_{tv}$ are set to $3.0$, $5.0$, and $0.02$, respectively.

\begin{figure*}[!htb]
	\centering
	\includegraphics[width=\textwidth]{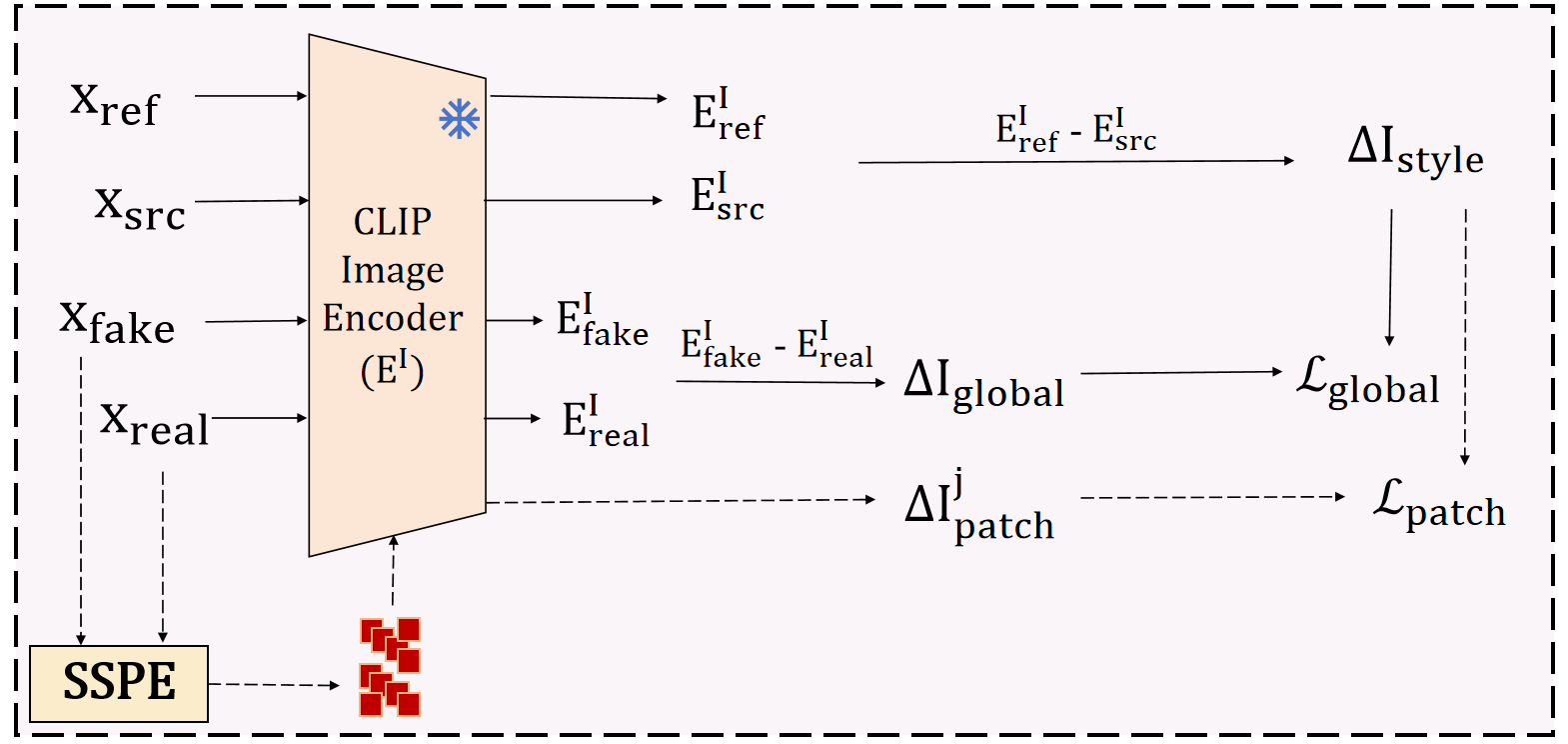}
	\caption{Illustration of the loss calculation for image-guided style transfer.}
	\label{fig:clip_image}
\end{figure*}

The above text-guided method uses the text direction vector $\Delta T$ as the target semantic direction. The directional supervision mechanism of CLIP-SSPE can be further extended to image-guided style transfer tasks. As shown in Fig. \ref{fig:clip_image}, the image-guided method uses the target style image to provide the change direction information.
Let $x_{src}$ denote the original image and $x_{ref}$ denote the target style image. FRPSS inputs the two images into the frozen CLIP image encoder $E^I$, respectively, to obtain the corresponding image features $E^{I}_{src}$ and $E^{I}_{ref}$. Subsequently, the style change direction vector $\Delta I_{style}$ is constructed according to the difference between the target image style feature vector and the original image feature vector:
\begin{equation}\label{eq:20}
	\Delta I_{style} = E^{I}_{ref} - E^{I}_{src}.
\end{equation}
During image-guided style transfer, $\Delta I_{style}$ replaces the text direction vector $\Delta T$ in Eq. \eqref{e11}, and the global and patch-wise directional constraints are still used for optimization. In this way, the CLIP-SSPE module uses the visual information provided by the target style image $x_{ref}$ to guide the generated result toward the target style.

\section{Experiments}
\subsection{Experimental datasets}
FRPSS is evaluated on three general single-image generation datasets to examine the generation capability for images with different textures and semantic structures, including Places50 \cite{Shaham2019SinGAN}, MSID16 \cite{Shaham2019SinGAN}, and SIGD16 \cite{granot2022gpnn}. Places50 contains 50 scene images with complex semantic structures. MSID16 contains 16 images covering multiple types of visual content, including plants, animals, and natural landscapes. SIGD16 contains 16 images with significant topological variations and is used to evaluate the generation stability of the model. The dataset adopted by SinDDM \cite{kulikov2023sinddm} is used for the text-guided stylization experiments. The dataset provided by InstantStyle-Plus \cite{wang2024instantstyle} is used for the image-guided stylization experiments.

\subsection{Evaluation metrics}
Single Image Fr\'echet Inception Distance (SIFID) \cite{Shaham2019SinGAN} and Learned Perceptual Image Patch Similarity (LPIPS) \cite{zhang2018unreasonable} are used to quantitatively evaluate the generated results.

SIFID follows the Fr\'echet distance form of the standard FID \cite{heusel2017gans}, but the statistical object is changed from the feature distribution of an image set to the spatial feature distribution within a single image. Given a real image $x_r$ and a generated image $x_f$, a shallow convolutional block of the pre-trained Inception-V3 \cite{szegedy2016rethinking} is used to extract the feature map $\psi(x) \in \mathbb{R}^{C \times H \times W}$. The feature map is then unfolded along the spatial dimensions into $H \times W$ $C$-dimensional feature vectors to estimate the mean $\mu \in \mathbb{R}^C$ and covariance matrix $\Sigma \in \mathbb{R}^{C \times C}$ of the local feature distribution of the image. The SIFID between the real image and a single generated image is defined as follows:
\begin{equation}\label{e21}
	\mathrm{SIFID}(x_r, x_f) = \|\mu_r - \mu_f\|_2^2 + \mathrm{Tr}\left(\Sigma_r + \Sigma_f - 2\left(\Sigma_r \Sigma_f\right)^{1/2}\right),
\end{equation}
where $\mu_r, \Sigma_r$ and $\mu_f, \Sigma_f$ denote the feature means and covariance matrices corresponding to the real image and the generated image, respectively, and $\mathrm{Tr}(\cdot)$ denotes the trace of a matrix. For each original image, $20$ samples are generated by the model, and the SIFID of each generated sample with respect to the original image is calculated. The mean SIFID is used as the final evaluation result. A lower SIFID value indicates that the distribution of the generated image is closer to that of the original image in terms of deep feature statistics.

To measure the diversity among the generated results, the pairwise LPIPS distances among the generated samples corresponding to the same original image are calculated. The LPIPS value is calculated based on the deep features extracted by the pre-trained AlexNet \cite{krizhevsky2012imagenet, Shaham2019SinGAN} network $\phi(x)$. For any two generated images $x$ and $y$, their LPIPS distance is defined as follows \cite{zhang2018unreasonable}:
\begin{equation}\label{e22}
	d(x,y) = \sum_{l=1}^{L} \frac{1}{H_l W_l} \sum_{h=1}^{H_l}\sum_{w=1}^{W_l} \left\| w_l \odot \left(\hat{\phi}_{l}(x)_{h,w} - \hat{\phi}_{l}(y)_{h,w}\right) \right\|_2^2,
\end{equation}
where $\phi_l(x) \in \mathbb{R}^{C_l \times H_l \times W_l}$ denotes the feature map output by the $l$-th layer of the pre-trained AlexNet, $H_l$ and $W_l$ denote the height and width of the feature map at the layer, $(h,w)$ denotes the spatial position coordinates on the feature map, $\phi_l(x)_{h,w} \in \mathbb{R}^{C_l}$ denotes the channel feature vector corresponding to the position, $\hat{\phi}_l(\cdot)$ denotes the feature obtained by normalizing $\phi_l(\cdot)$ along the channel dimension, and $w_l$ denotes the linear weighting coefficient of the $l$-th layer. For the $20$ samples generated from the same original image, the LPIPS distances of all $190$ non-repeated image pairs are calculated, and their mean is taken as the diversity metric. A higher LPIPS value indicates a larger perceptual difference among the generated samples, while a lower value indicates a higher degree of homogeneity among the samples.

\subsection{Qualitative and quantitative analysis of image generation}
\begin{figure*}[!htb] 
	\centering
	\includegraphics[width=\textwidth]{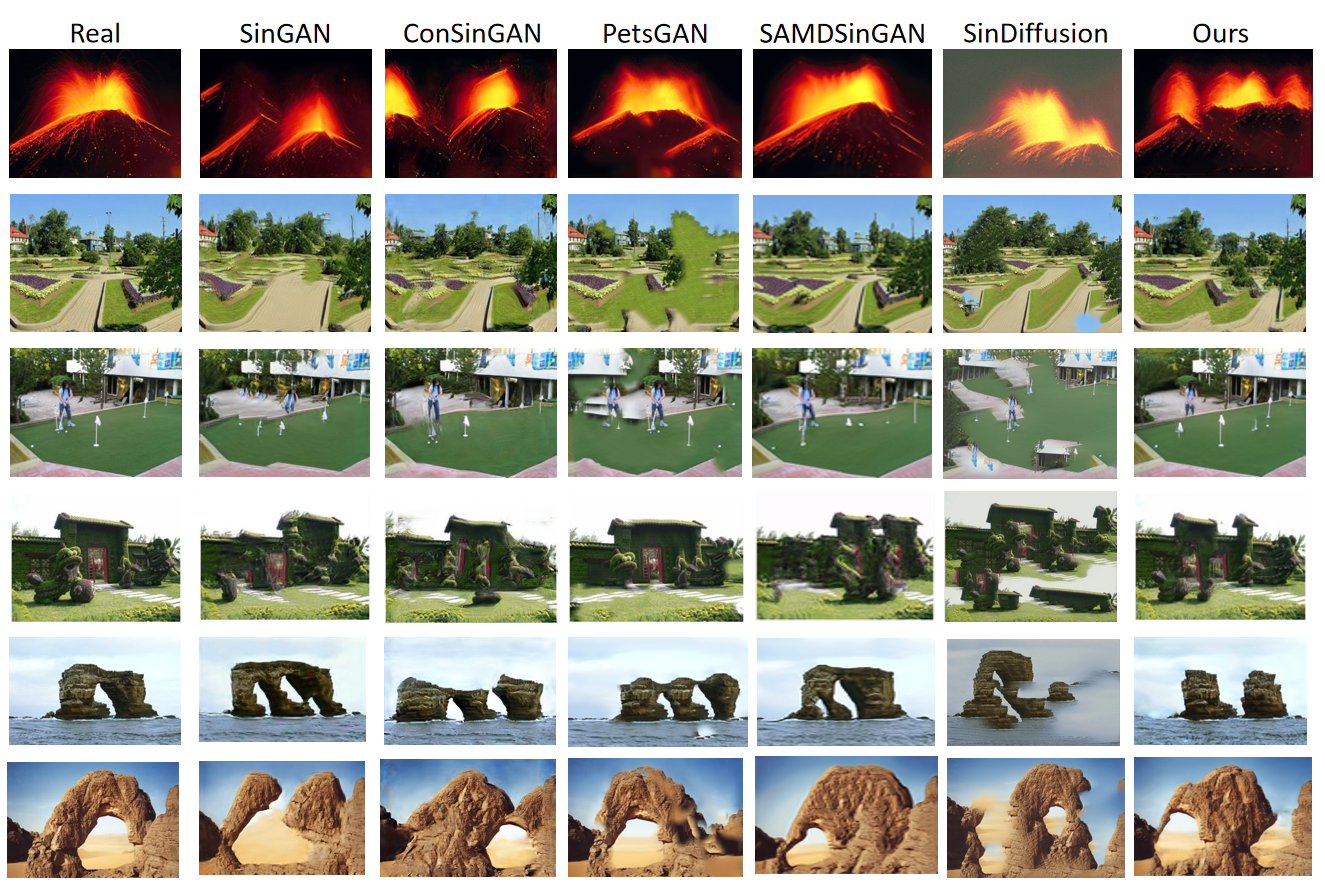} 
	\caption{Images generated by different methods on the Places50 dataset.}
	\label{fig:places50} 
\end{figure*}

\begin{figure*}[!htb] 
\centering
\includegraphics[width=\textwidth]{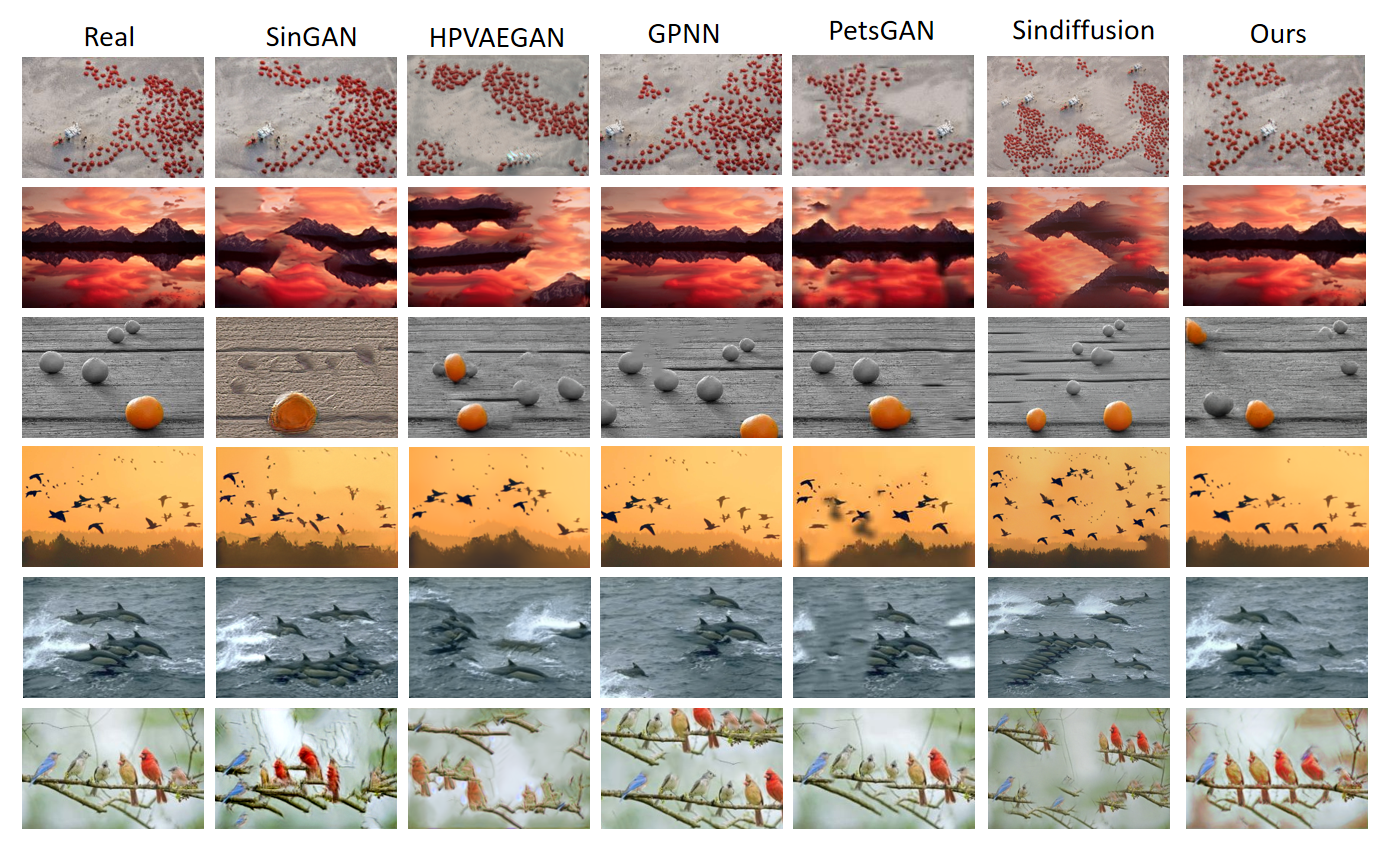} 
\caption{Images generated by different methods on the SIGD16 dataset.}
\label{fig:sigd16} 
\end{figure*}

Fig. \ref{fig:places50} and Fig. \ref{fig:sigd16} visually compare the generation results of FRPSS with those of other methods on the Places50 and SIGD16 datasets.

Some results generated by traditional multi-scale generation models, such as SinGAN \cite{Shaham2019SinGAN} and ConSinGAN \cite{Hinz_2021WACV_cosingan}, show obvious structural misalignment. Structural deviations produced at low-resolution stages may accumulate during the progressive generation process, further leading to local incoherence such as path distortion and mountain discontinuity. Similarly, although HP-VAE-GAN \cite{NEURIPS2020_hpvaegan} achieves a relatively high LPIPS score, obvious global structural misalignment can still be observed in some generated results. GPNN \cite{granot2022gpnn} can synthesize relatively realistic local textures, but the method tends to copy the nearest image patches from the training image, thereby limiting the diversity of generated samples to some extent. In addition, diffusion-based single-image generation methods such as SinDiffusion \cite{wang2025sindiffusion} have a strong capability for detail synthesis, but color leakage between regions or local stitching incoherence can still be observed in some generated results.

Compared with the above methods, FRPSS achieves a better balance between global structural integrity and local visual fidelity. MSR-FAGS performs feature rearrangement and matching in the low-scale feature space to provide more possibilities for global layout variations, while preserving the coherence of the spatial structure as much as possible. As shown in the figures, the images generated by FRPSS exhibit fewer spatial misalignment artifacts and maintain better consistency with the original image in terms of color distribution and contrast.

Table~\ref{tab:t1} reports the quantitative comparison results on the Places50, MSID16, and SIGD16 datasets. In terms of SIFID, FRPSS achieves the best results on all three datasets, indicating that the generated results of FRPSS are closer to the original images in terms of deep feature statistical distributions. In contrast, some comparison methods, such as SinDDM and SinDiffusion, obtain relatively high LPIPS scores, but their SIFID values are also relatively high, indicating larger deviations between the generated results and the feature distributions of the original images. Methods such as ConSinGAN and SinGAN have relatively low SIFID values, but their LPIPS scores are also relatively low, indicating limited diversity. The results show that FRPSS can still maintain local texture statistical characteristics similar to those of the original images while introducing feature variations.
\begin{table*}[!htb]
	\centering
	\caption{Quantitative comparison on the three datasets.}
	\label{tab:t1}
	\footnotesize  
	\setlength{\tabcolsep}{2.0pt}  
	\renewcommand{\arraystretch}{1.4}  
	\begin{tabular*}{\linewidth}{@{\extracolsep{\fill}} l cc cc cc @{}}
		\toprule
		
		& \multicolumn{2}{c}{Places50}
		& \multicolumn{2}{c}{MSID16}
		& \multicolumn{2}{c}{SIGD16} \\
		\cmidrule(lr){2-3} \cmidrule(lr){4-5} \cmidrule(lr){6-7} 
		& SIFID$\downarrow$ & LPIPS$\uparrow$ 
		& SIFID$\downarrow$ & LPIPS$\uparrow$ 
		& SIFID$\downarrow$ & LPIPS$\uparrow$ \\
		\midrule
		SinGAN        & 0.09    & 0.266   
		& 0.28    & 0.279   
		& 0.27    & 0.344   \\
		HP-VAE-GAN    & 0.30    & \underline{0.380}   
		& \underline{0.05}    & \underline{0.447}      
		& 0.07    & \textbf{0.449}   \\
		ConSinGAN     & 0.06   & 0.305 
		& 0.28 & 0.296 
		& \underline{0.06} & 0.287 \\
		GPNN          & 0.07 & 0.238 
		& 0.57 & 0.289 
		& 0.11 & 0.317 \\
		SinDDM        & 0.68 & 0.345 
		& 0.36 & 0.364 
		& 0.85 & 0.349 \\
		PetsGAN       & 0.08 & 0.310 
		& 0.29 & 0.342 
		& 0.32 & 0.340 \\
		SinFusion     & 0.64 & 0.368 
		& 0.48 & 0.339 
		& 0.69 & 0.365 \\
		MultiOSG      & 0.39 & 0.310   
		& 0.20 & 0.382   
		& 0.13 & 0.362  \\
		
		SinDiffusion  & 0.06 & \textbf{0.387} 
		& 0.49 & \textbf{0.489} 
		& 0.43 & \underline{0.445} \\
		StructDiff    & \underline{0.04} & 0.311
		& 0.40   & 0.368    
		& 0.14 & 0.330    \\
		Ours 		  & \textbf{0.02} & 0.321 
		& \textbf{0.03} & 0.386 
		& \textbf{0.03} & 0.356 \\
		\bottomrule
		
	\end{tabular*}
	\vspace{0.3em}
	\scriptsize 
	\begin{tabular}{@{}l@{}}
		Note: SIFID ($\downarrow$) means lower is better; LPIPS ($\uparrow$) means higher is better.\\
		Boldface indicates the best result; underlining indicates the second-best result.
	\end{tabular}
\end{table*}

In terms of generation diversity, FRPSS achieves competitive LPIPS scores on all three datasets, indicating that the generated samples maintain good perceptual diversity while preserving the visual characteristics of the original images.
\subsection{Image stylization transformation}
The section presents the applications of FRPSS to multiple downstream image manipulation tasks, including text-guided image style transfer, text-guided image content generation, image-guided style transfer, and paint-to-image.

\begin{figure*}[!htb] 
	\centering
	\includegraphics[width=\textwidth]{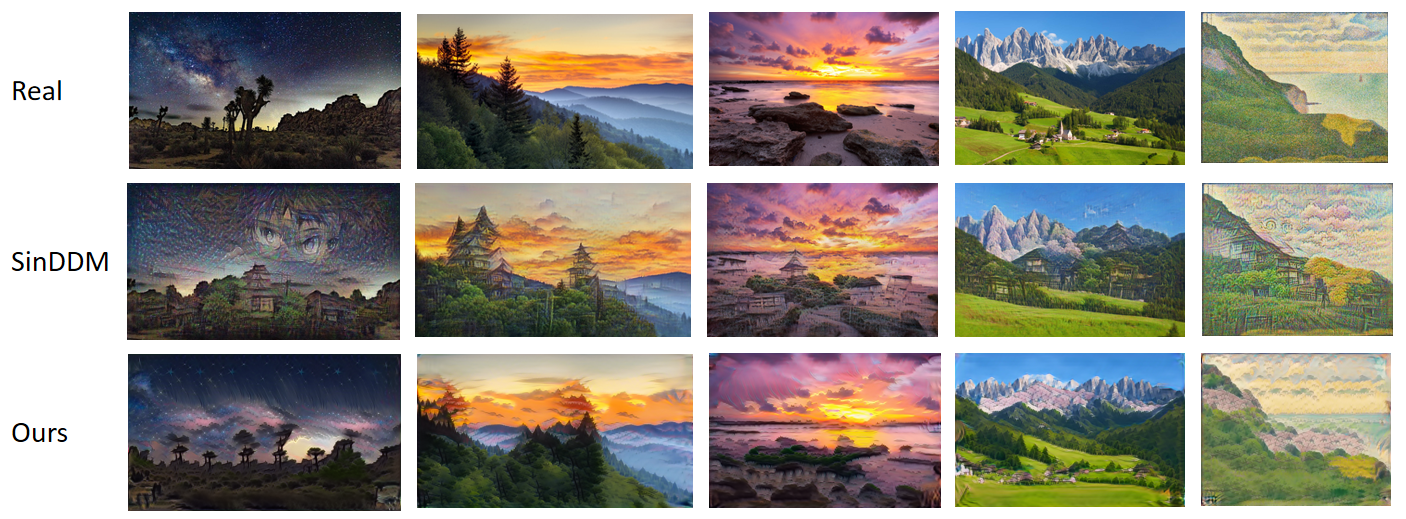} 
	\caption{Text-guided stylization results using anime-style text prompts.}
	\label{fig:tgst_anime} 
\end{figure*}

\begin{figure*}[!htb] 
	\centering
	\includegraphics[width=\textwidth]{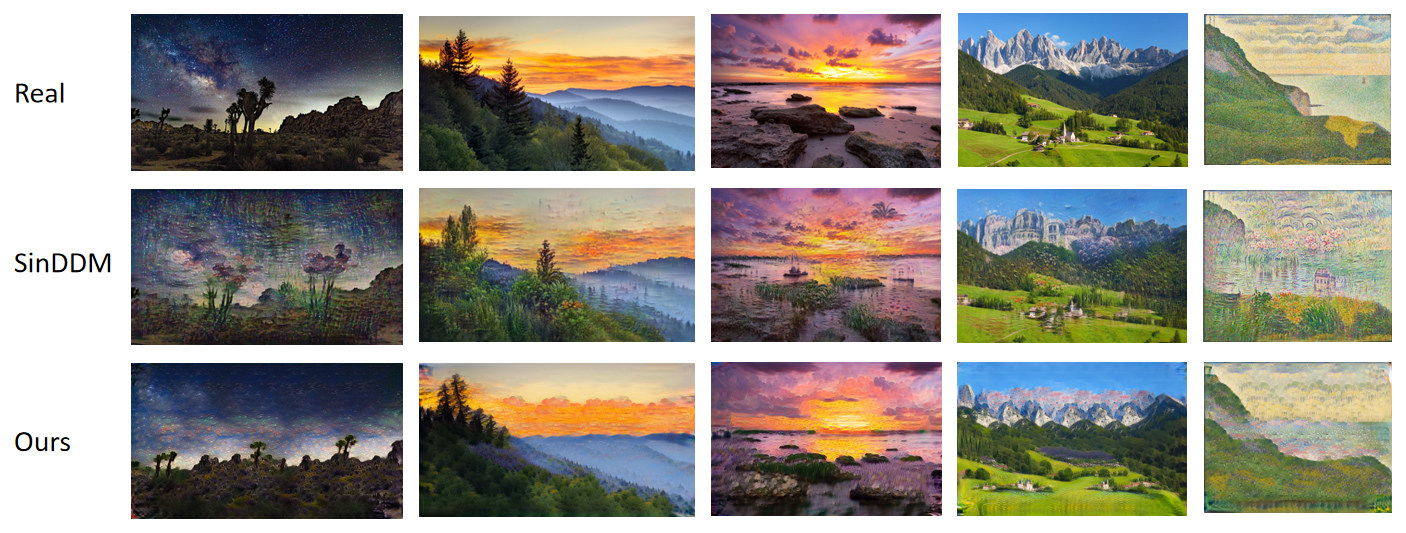} 
	\caption{Text-guided stylization results using Monet-style text prompts.}
	\label{fig:tgst_monet} 
\end{figure*}

\begin{figure*}[!htb] 
	\centering
	\includegraphics[width=\textwidth]{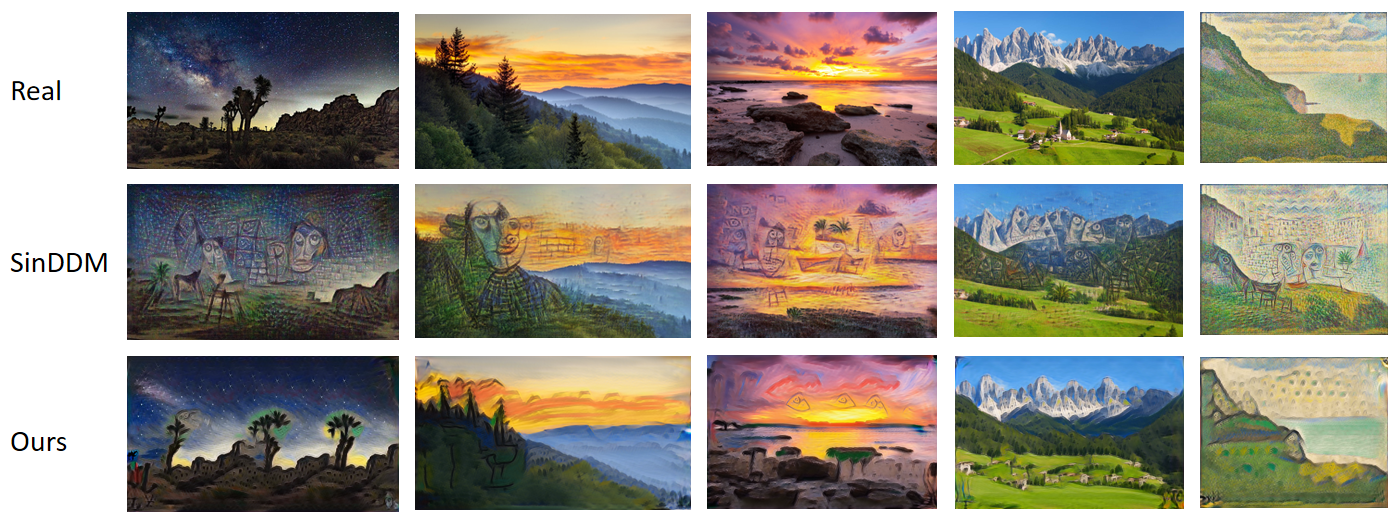} 
	\caption{Text-guided stylization results using Picasso-style text prompts.}
	\label{fig:tgst_picasso} 
\end{figure*}

\begin{figure*}[!htb] 
	\centering
	\includegraphics[width=\textwidth]{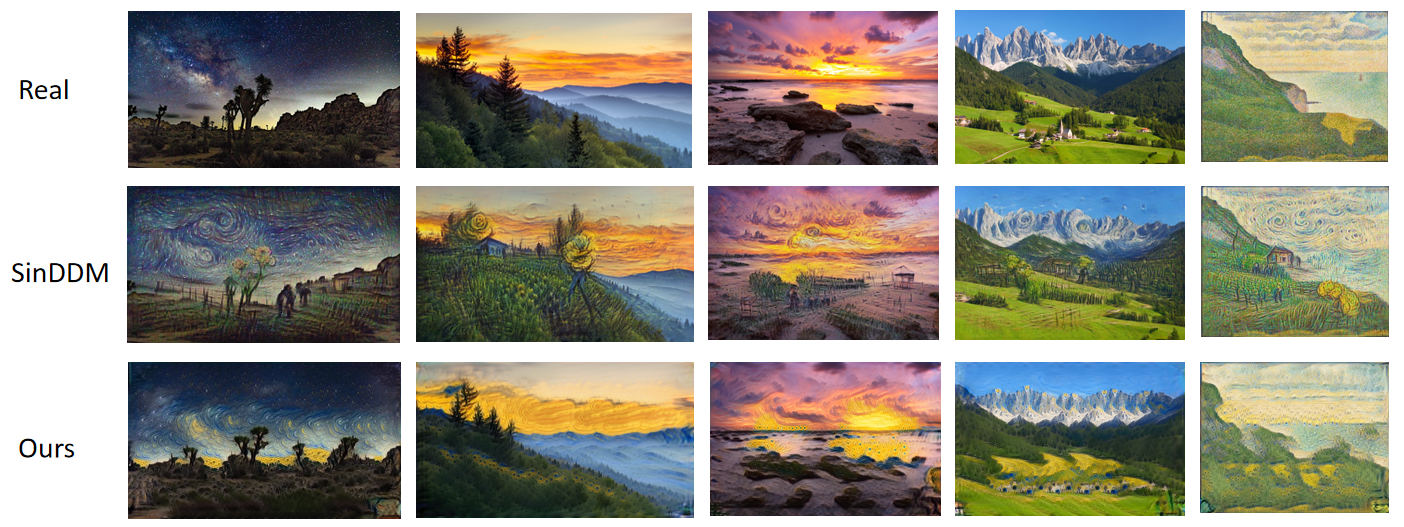} 
	\caption{Text-guided stylization results using Van Gogh-style text prompts.}
	\label{fig:tgst_vg} 
\end{figure*}

\textbf{Text-guided style transfer.}
Here, $src$ is set to ``a photo'', and $trg$ is set to ``anime style'', ``Monet style'', ``Picasso style'', and ``Van Gogh style''. CLIP-SSPE constructs directional constraints according to the visual change direction from the real image to the generated image and the semantic change direction from the source text to the target text, so that the generated results preserve the main content and spatial structure of the original image as much as possible while introducing the target style. Fig. \ref{fig:tgst_anime}, Fig. \ref{fig:tgst_monet}, Fig. \ref{fig:tgst_picasso}, and Fig. \ref{fig:tgst_vg} compare the generated results of FRPSS and SinDDM \cite{kulikov2023sinddm} under the four text prompts.

From the generated results, SinDDM has difficulty in completely preserving the original scene content during some text-guided style transfer processes. SinDDM is a diffusion-based single-image generation method and performs semantic guidance during the iterative denoising sampling process. Therefore, some generated results may exhibit local content changes together with changes in visual style. For example, in Fig. \ref{fig:tgst_anime} and Fig. \ref{fig:tgst_picasso}, when $trg$ is set to ``Anime Style'' and ``Picasso Style'', faces or abstract elements that do not exist in the original image are introduced into the sky or mountain regions in some results. In Fig. \ref{fig:tgst_monet} and Fig. \ref{fig:tgst_vg}, when $trg$ is set to ``Van Gogh Style'' and ``Monet Style'', local semantic content changes can also be observed. The above phenomena indicate that SinDDM may introduce content that does not exist in the original image during text-guided style transfer. In contrast, FRPSS can better preserve the main content of the original image, such as Joshua trees, grasslands, and rocks, while reducing the generation of additional content.

\begin{figure*}[!htb] 
	\centering
	\includegraphics[width=\textwidth]{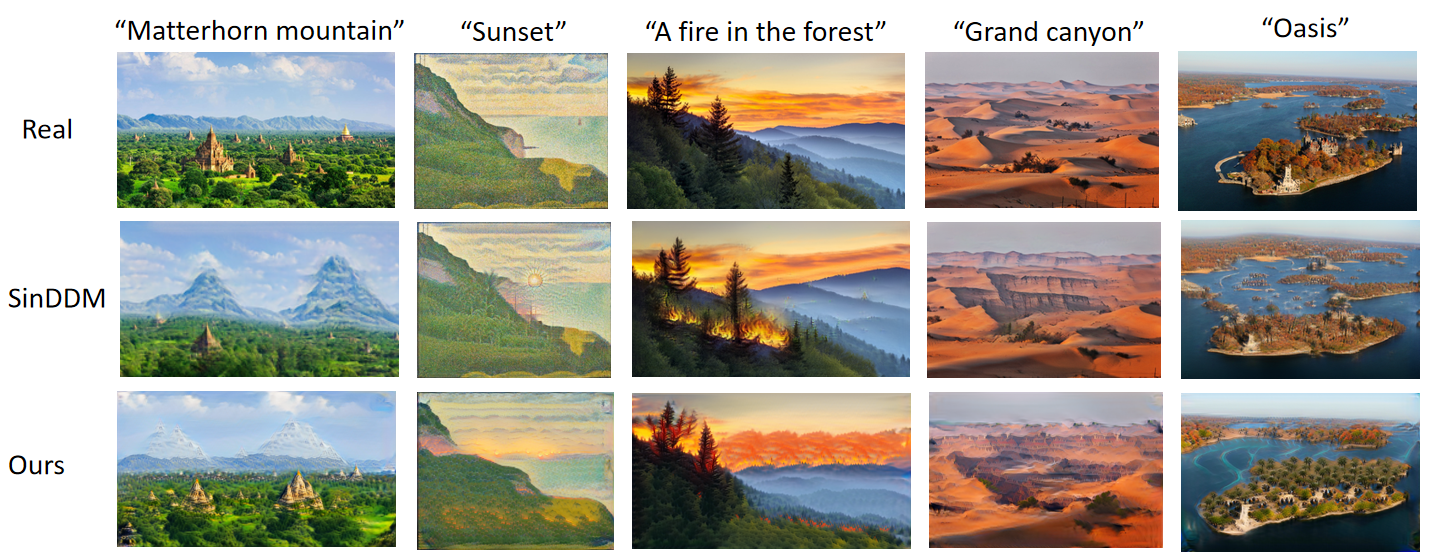} 
	\caption{Text-guided image content generation results.}
	\label{fig:tgcg_1} 
\end{figure*}

\textbf{Text-guided content generation.}
The text-guided image content generation task requires the model not only to change the visual appearance of the image, but also to generate content consistent with the target semantics according to the text prompt. The same input form as text-guided style transfer is adopted for the task. $src$ is still set to ``a photo'', while $trg$ is set according to the desired semantic content, including ``Matterhorn mountain'', ``Sunset'', ``A fire in the forest'', ``Grand canyon'', and ``Oasis''. Fig. \ref{fig:tgcg_1} shows the generated results of FRPSS and the comparison method under different text prompts.
When $trg$ is set to ``Oasis'', SinDDM fails to form an obvious oasis landscape. In addition, when $trg$ is set to ``A fire in the forest'' and ``Sunset'', some generated results of SinDDM mainly show changes in color or local texture and still differ from the semantic content described by the target text. In contrast, FRPSS can produce more obvious target semantic changes.

\begin{figure*}[!htb] 
	\centering
	\includegraphics[width=\textwidth]{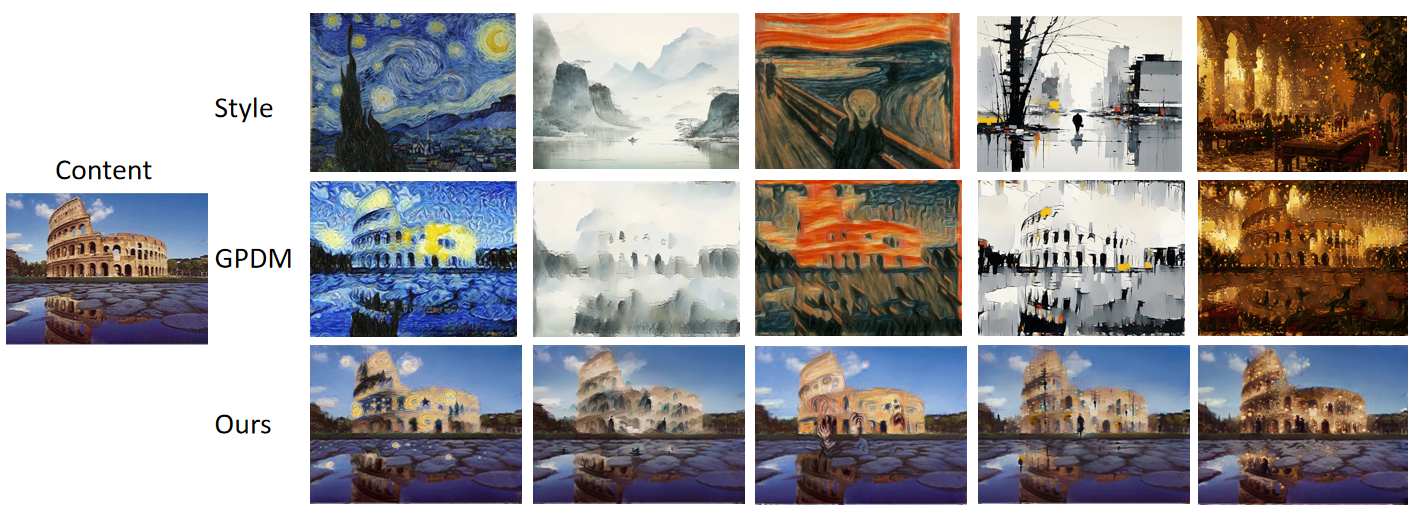} 
	\caption{Image-guided style transfer results.}
	\label{fig:igst} 
\end{figure*}

\textbf{Image-guided style transfer.}
The performance of FRPSS on the image-guided style transfer task is further evaluated. Here, $x_{src}$ is set to the content image shown on the left side of Fig. \ref{fig:igst}, and $x_{ref}$ is set to the five style reference images shown in the first row on the right side of Fig. \ref{fig:igst}, including Van Gogh style, ink landscape painting style, expressionist style, watercolor city style, and golden hall style. The second and third rows on the right side of Fig. \ref{fig:igst} show the generated results of GPDM \cite{elnekave2022generating} and FRPSS on the task.

As shown in Fig. \ref{fig:igst}, GPDM tends to produce over-smoothing during image-level feature injection, which makes the boundaries of some main structures in the original image blurred. In contrast, FRPSS can better transfer the texture features from the target style image to the original image while preserving the clarity of the main structures in the original image.

\begin{figure*}[!htb] 
	\centering
	\includegraphics[width=\textwidth]{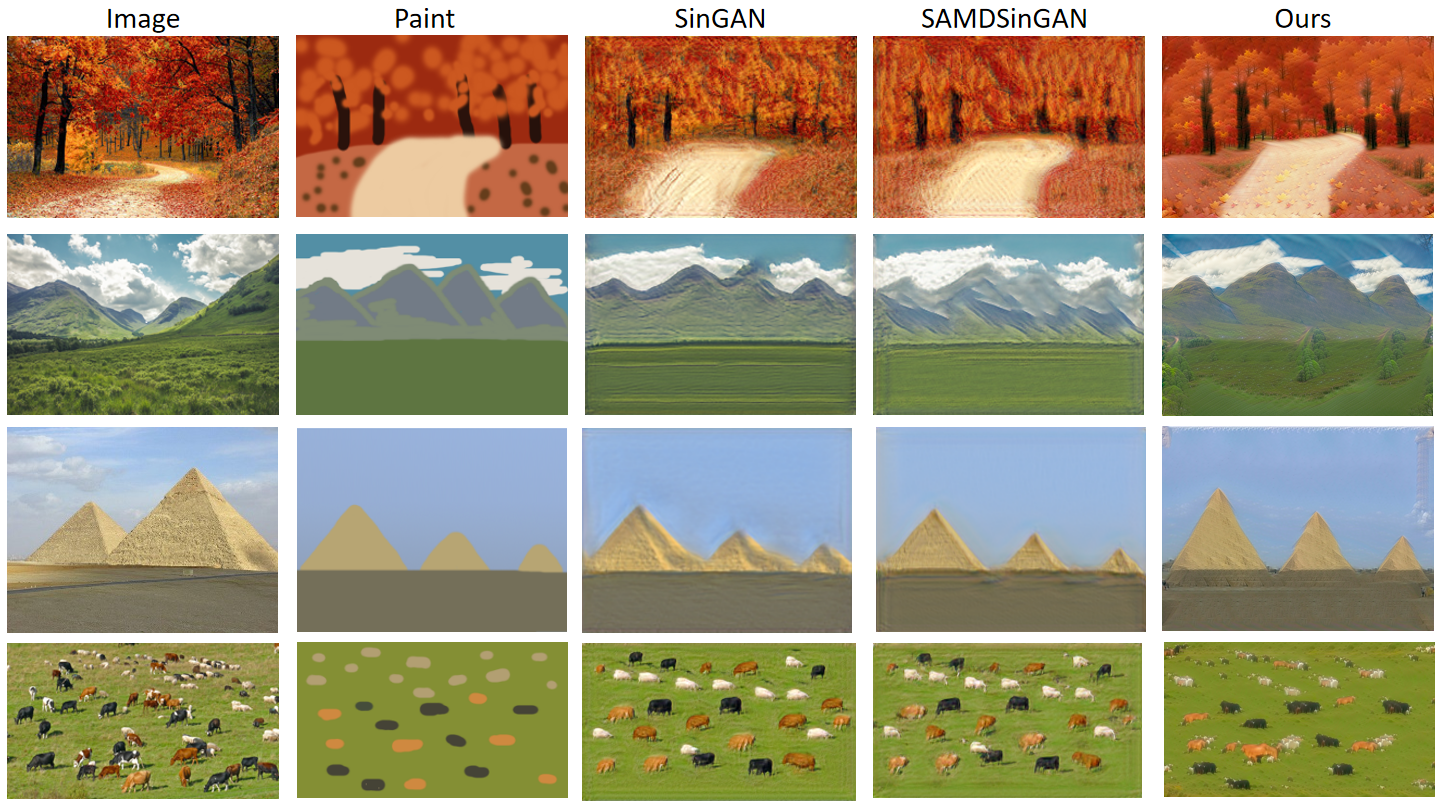} 
	\caption{Paint-to-image results.}
	\label{fig:pti} 
\end{figure*}

\textbf{Paint-to-image.}
The goal of the paint-to-image task is to generate images with natural textures and visual details according to the structural layout provided by a paint. FRPSS follows the directional guidance mechanism used in image-guided style transfer. Here, $x_{src}$ is set to the paint shown in the Paint column of Fig. \ref{fig:pti}, and $x_{ref}$ is set to the natural image shown in the Image column.

Baseline methods such as SinGAN and SAMDSinGAN usually inject paint information at specific low-scale stages of the generation process, but the implicit feature fusion tends to make the texture transfer in the generated results relatively rigid. In contrast, FRPSS can progressively add textures and visual details while preserving the main spatial layout of the paint, thereby generating more natural image results.

\subsection{Outpainting}
\begin{figure*}[!htb]
	\centering
	\includegraphics[width=\textwidth]{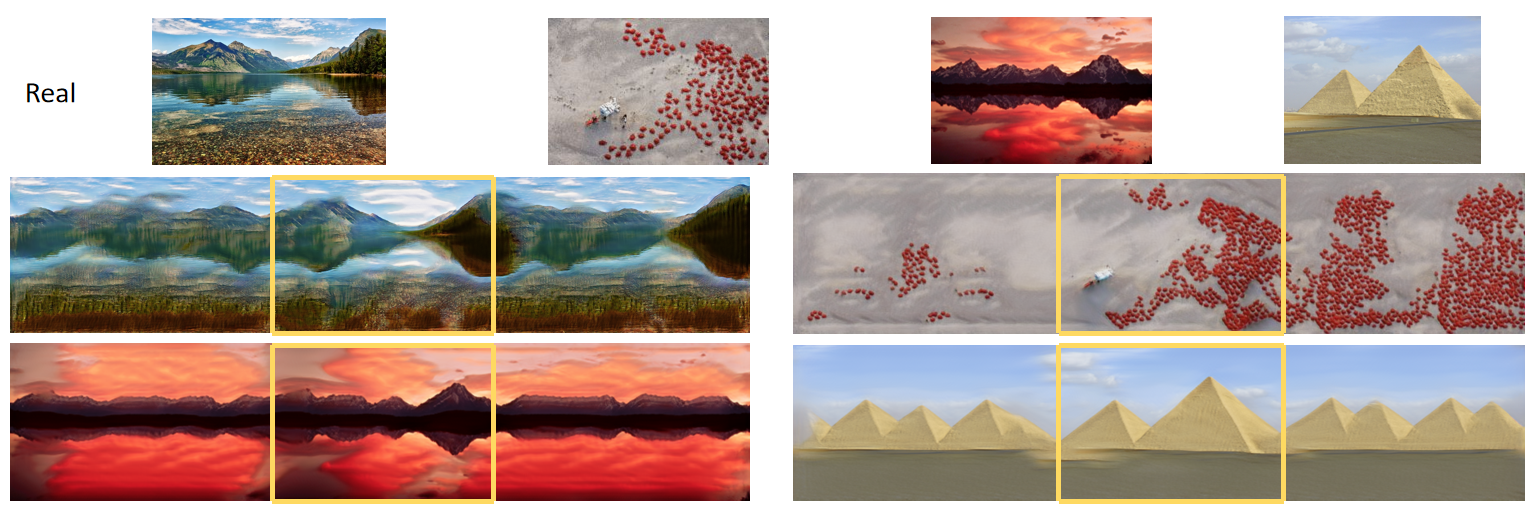}
	\caption{Image outpainting. The yellow boxes indicate the original images.}
	\label{fig:outpainting} 
\end{figure*}

The goal of the outpainting task is to generate visual content beyond the boundaries of the original image. Following the implementation of MultiOSG \cite{gou2025multiosg}, during inference, FRPSS first encodes the low-scale image to obtain the feature vector $Z$ of the original image. Subsequently, the MSR-FAGS module generates the left extension feature $Z^{*}_{left}$ and the right extension feature $Z^{*}_{right}$, respectively. The two extension features are then concatenated with the original feature $Z$ along the width $W$. Since the generator has a fully convolutional structure, its parameters do not depend on a fixed spatial size of the input feature. Therefore, the concatenated extended feature can be directly input into the generator for subsequent generation. During the subsequent multi-scale generation process, bilinear interpolation is still applied to the current outpainting result at each scale according to the original scale ratio to progressively adjust its spatial resolution. As shown in Fig. \ref{fig:outpainting}, the generated outpainting regions maintain good continuity with the original image in terms of visual texture and overall scene layout.

\subsection{Parameter influence analysis}
\begin{figure*}[!htb]
	\centering
	\includegraphics[width=\linewidth]{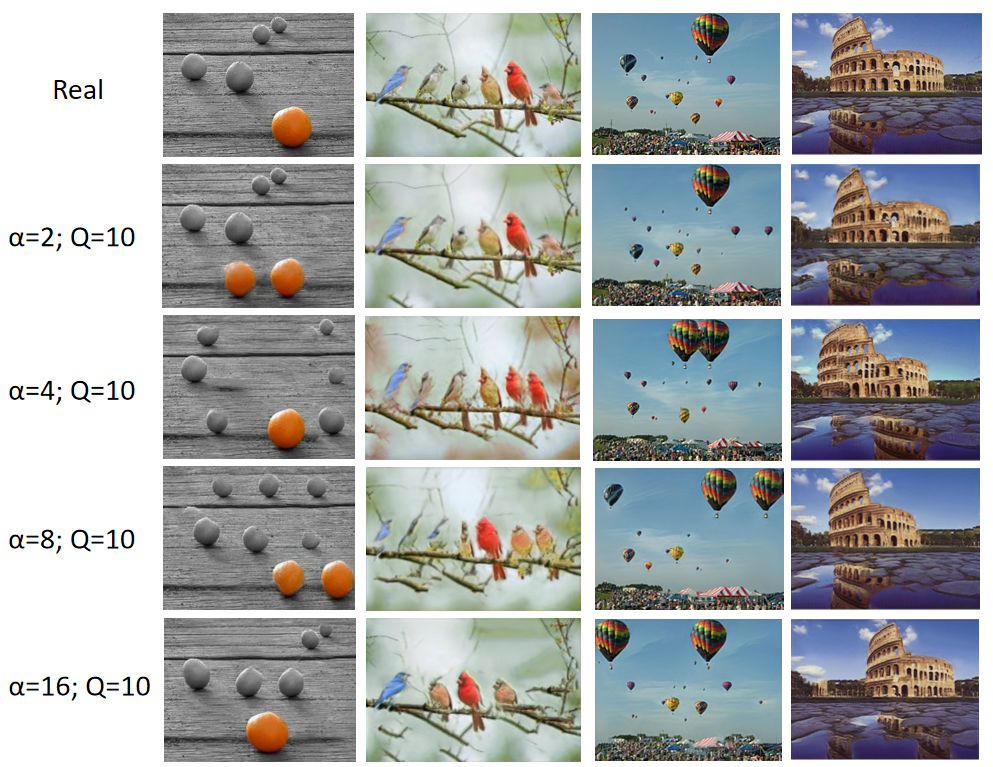}
	\caption{Effect of the feature slice width $\alpha$.}
	\label{fig:ablation_alpha}
\end{figure*}

\begin{figure*}[!htb]
	\centering
	\includegraphics[width=\linewidth]{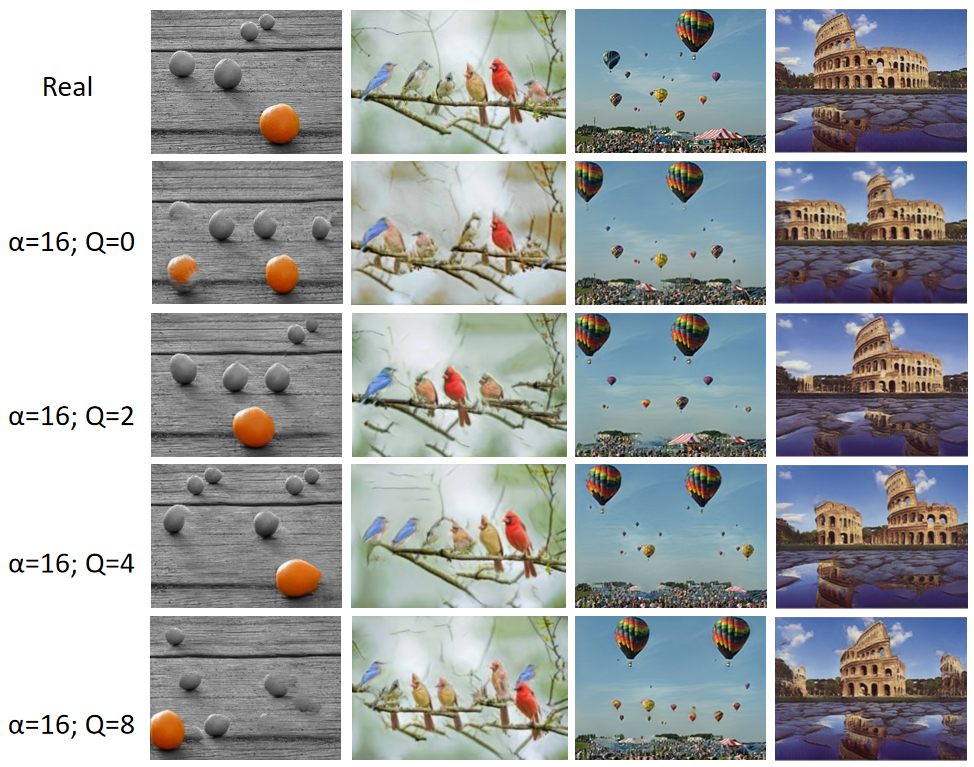} 
	\caption{Effect of the number of interpolated features $Q$. The red boxes show bird images with obvious variations generated by FRPSS.}
	\label{fig:ablation_gamma}
\end{figure*}

The key hyperparameters of the MSR-FAGS module are quantitatively and qualitatively analyzed on the SIGD16 dataset, with a focus on the effects of the feature slice width $\alpha$ and the number of interpolated features $Q$ on generation quality and diversity. The quantitative results are shown in Table \ref{tab:ablation_params}, and the corresponding qualitative results are shown in Fig. \ref{fig:ablation_alpha} and Fig. \ref{fig:ablation_gamma}.
\begin{table}[!htb]
	\centering
	\caption{Parameter influence analysis of the MSR-FAGS module on the SIGD16 dataset. SIFID ($\downarrow$) and LPIPS ($\uparrow$) results under different feature slice widths ($\alpha$) and numbers of interpolated features ($Q$).}
	\label{tab:ablation_params}
	\begin{tabular*}{\linewidth}{@{\extracolsep{\fill}}lccc@{}}
		\toprule
		Settings & Configuration & SIFID $\downarrow$ & LPIPS $\uparrow$ \\
		\midrule
		\multirow{4}{*}{\shortstack{Varying $\alpha$ \\ ($Q=2$)}} 
		& $\alpha=2$  & 0.043 & 0.125 \\
		& $\alpha=4$  & 0.034 & 0.244 \\
		& $\alpha=8$  & \textbf{0.031} & 0.317 \\
		& $\alpha=16$ & 0.033 & \textbf{0.353} \\
		\midrule
		\multirow{4}{*}{\shortstack{Varying $Q$ \\ ($\alpha=16$)}} 
		& $Q=0$   & 0.033 & \textbf{0.356} \\
		& $Q=2$ & 0.032 & 0.353 \\
		& $Q=4$   & 0.032 & 0.349 \\
		& $Q=8$  & \textbf{0.031} & 0.347 \\
		\bottomrule
	\end{tabular*}
\end{table}

Table \ref{tab:ablation_params} shows the effects of the feature slice width $\alpha$ on generation fidelity and diversity. When $\alpha=2$, the LPIPS value of the model decreases to 0.125, and the SIFID is 0.043. Combined with the qualitative results in Fig. \ref{fig:ablation_alpha}, an excessively small $\alpha$ limits the spatial rearrangement range of the features and makes the model tend to reproduce the original image, thereby limiting the diversity of the generated results. As $\alpha$ increases to 16, LPIPS increases to 0.353, while SIFID remains at a relatively low level (0.033). The result indicates that appropriately increasing $\alpha$ can improve the diversity of the generated samples while keeping SIFID at a relatively low level.

With $\alpha=16$ fixed, the effect of the number of interpolated features $Q$ is further examined. When $Q=0$, the model achieves the highest LPIPS score (0.356), but the generated results tend to show local stitching artifacts due to the lack of intermediate features generated by interpolation. As the number of interpolated features $Q$ increases to 8, LPIPS slightly decreases to 0.347, while SIFID further decreases to 0.031. The red regions in Fig. \ref{fig:ablation_gamma} indicate that introducing an appropriate amount of feature interpolation helps the model synthesize more natural and coherent semantic features, thereby generating images with both structural plausibility and novelty.

\subsection{Stylization ablation experiment}
\begin{figure*}[!htb]
	\centering
	\includegraphics[width=\linewidth]{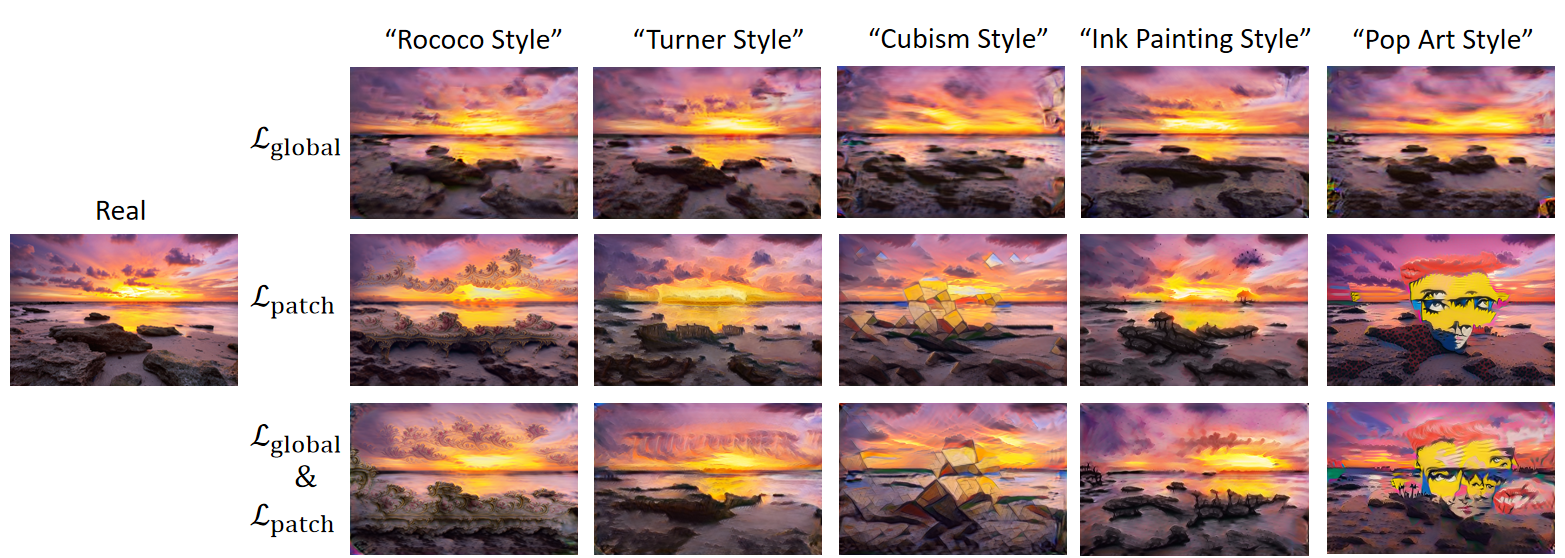} 
	\caption{Effects of the global and patch-wise CLIP losses on the stylization results.}
	\label{fig:ablation_style}
\end{figure*}

The effects of the global directional loss $\mathcal{L}_{global}$ and the local directional loss $\mathcal{L}_{patch}$ on the text-guided stylization results are analyzed. Here, the original image is the sunset scene image shown on the left side of Fig. \ref{fig:ablation_style}. $src$ is set to ``a photo'', while $trg$ is set to ``Rococo Style'', ``Turner Style'', ``Cubism Style'', ``Ink Painting Style'', and ``Pop Art Style'', respectively. As shown in Fig. \ref{fig:ablation_style}, when only $\mathcal{L}_{global}$ is used, the generated results mainly show changes in the overall color tone, while the local style textures are relatively weak. When only $\mathcal{L}_{patch}$ is used, more obvious local textures can be introduced into the generated results, but the overall visual consistency is relatively insufficient. When the two losses are jointly used, both global style changes and local texture generation can be taken into account.

\subsection{Limitations}
\begin{figure*}[!htb]
	\centering
	\includegraphics[width=\textwidth]{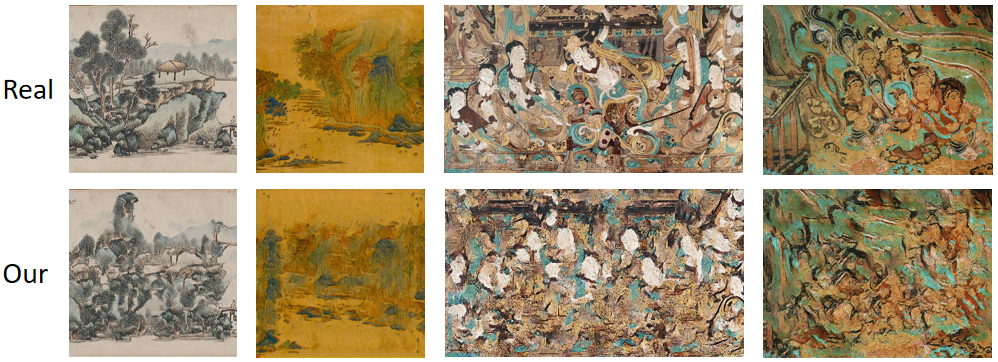}
	\caption{Failure cases of FRPSS.}
	\label{fig:fail}
\end{figure*}

Although FRPSS achieves good generation results on various single-image generation tasks, it still has certain limitations in some complex scenes. As shown in Fig. \ref{fig:fail}, for images such as Chinese landscape paintings and murals that contain large background regions and sparse semantic objects, the generated results may exhibit local content missing. For example, small semantic regions such as thatched cottages and human faces in the original image are not completely preserved in some generated results. The phenomenon may be related to the structural rearrangement of MSR-FAGS in the low-scale feature space. In low-scale representations, large background regions occupy a relatively high spatial proportion, while local objects such as thatched cottages and human faces are represented by only a small number of feature regions. During feature rearrangement and matching, the sparse semantic regions may be affected by background features with larger proportions, thereby weakening their structural information and causing local content missing. In addition, the current stylization mechanism of FRPSS directly introduces the target semantic direction into the training process of the multi-scale generative model, resulting in a certain degree of coupling between the generative model and the specific target style. When the target style or guidance condition changes, the model usually needs to be retrained. Therefore, switching among multiple styles still incurs a certain training overhead.

\section{Conclusion}
Existing single-image generation methods usually lack explicit global structural constraints and are prone to structural misalignment or incoherent results during generation. To address the problem, the paper proposes the FRPSS single-image generation framework. FRPSS uses the MSR-FAGS module to rearrange, augment, and match the latent features of a single training image in the Pre-Shape Space, providing latent features with structural variations for the subsequent multi-scale generation process and thereby reducing the risk of structural misalignment during generation. In addition, the CLIP-SSPE module achieves controllable semantic guidance through global and patch-wise directional CLIP constraints. Experimental results show that FRPSS achieves good generation fidelity on three single-image datasets and realizes an effective trade-off between fidelity and diversity. Qualitative experiments further verify its effectiveness on downstream tasks with CLIP-SSPE, including text-guided style transfer, text-guided content generation, image-guided style transfer, and paint-to-image.

FRPSS still has certain limitations. On the one hand, in complex scenes containing large background regions and sparse semantic objects, low-scale feature rearrangement may weaken the structural information of local semantic regions. On the other hand, the current stylization mechanism directly introduces the target semantic direction into the training process of the multi-scale model, resulting in a certain degree of coupling between the generative model and the target style. When the target style or guidance condition changes, the model usually needs to be retrained, thereby limiting the flexibility and application efficiency of the model in scenarios requiring rapid switching among multiple styles. In the future, region importance modeling or attention mechanisms can be further introduced into MSR-FAGS to enhance the preservation of sparse semantic regions. Meanwhile, conditional injection mechanisms that decouple semantic control from the training process of the generative model can be explored, allowing the model to perform style control according to different texts or reference images during inference and thereby reducing the repeated training overhead for different target styles.

\clearpage

\printbibliography

\end{CJK}
\end{document}